\pdfoutput=1
\documentclass[11pt]{article}

\usepackage[]{EMNLP2022}
\usepackage{physics,amsmath}

\usepackage{times}
\usepackage{latexsym}

\usepackage[T1]{fontenc}
\usepackage[utf8]{inputenc}

\usepackage{microtype}

\usepackage{graphicx}
\usepackage{dsfont}
\usepackage{microtype}
\usepackage{amssymb}
\usepackage{multirow}

\usepackage{balance}
\usepackage{booktabs}
\usepackage[normalem]{ulem}
\usepackage{enumitem}
\usepackage{subcaption}
\usepackage{inconsolata}

\setlist[itemize,enumerate]{leftmargin=*}

\definecolor{royalblue}{rgb}{0.25, 0.41, 0.88}
\newcommand{\edit}[1]{\textcolor{black}{#1}}

\title{Disentangling Uncertainty in Machine Translation Evaluation}

\author{
Chrysoula Zerva$^{1, 4}$  %\and
Taisiya Glushkova$^{1, 4}$ %\and %\quad  
Ricardo Rei$^{2, 3, 4}$  %\and
André F. T. Martins$^{1, 2, 4}$ 
\\
$^1$Instituto de Telecomunicações \quad $^2$Unbabel \quad $^3$INESC-ID \\
$^4$Instituto Superior Técnico \& LUMLIS (Lisbon ELLIS Unit) \\
{\small \texttt{\{chrysoula.zerva, taisiya.glushkova, ricardo.rei, andre.t.martins\}@tecnico.ulisboa.pt}}\\
}

\begin{document}
\maketitle
\begin{abstract}
Trainable evaluation metrics for machine translation (MT) exhibit strong correlation with human judgements, but they are often hard to interpret and might produce unreliable scores under noisy or out-of-domain data. 
Recent work has attempted to mitigate this with simple uncertainty quantification techniques (Monte Carlo dropout and deep ensembles), 
however these techniques (as we show) are limited in several ways – for example, they are unable to distinguish between different kinds of uncertainty, and they are time and memory consuming. 
In this paper, we propose more powerful and efficient uncertainty predictors for MT evaluation, and we assess their ability to target different sources of aleatoric and epistemic uncertainty. 
To this end, we develop and compare training objectives for the \textsc{Comet} metric to enhance it with an uncertainty prediction output, including heteroscedastic regression, divergence minimization, and direct uncertainty prediction.
Our experiments show improved results on uncertainty prediction for the WMT metrics task datasets, with a substantial reduction in computational costs. Moreover, they demonstrate the ability of these predictors to address specific uncertainty causes in MT evaluation, such as low quality references and out-of-domain data.\footnote{Our code and data is available at: \url{https://github.com/deep-spin/uncertainties_MT_eval}}% \url{https://anonymous.4open.science/r/uncertainties_MT_eval-C2DB}}
%\andre{We could say here that we release our code?}

%Despite recent progress in machine translation evaluation, MT evaluation systems have rarely provided uncertainty estimates regarding their quality outputs. And even when they have, it was hard to tell what kind of uncertainty the proposed methods are handling and how well.

%In this paper we are looking into more powerful techniques of uncertainty-focused MT evaluation that overcome these limitations. We explore variations of such approaches as direct uncertainty prediction and heteroscedastic regression, and show that they outperform previously considered ones, like MCD and deep ensembles. 

\end{abstract}

\section{Introduction}
\label{sec:intro}

\begin{figure}[t]
\centering
\includegraphics[width=0.9\columnwidth] {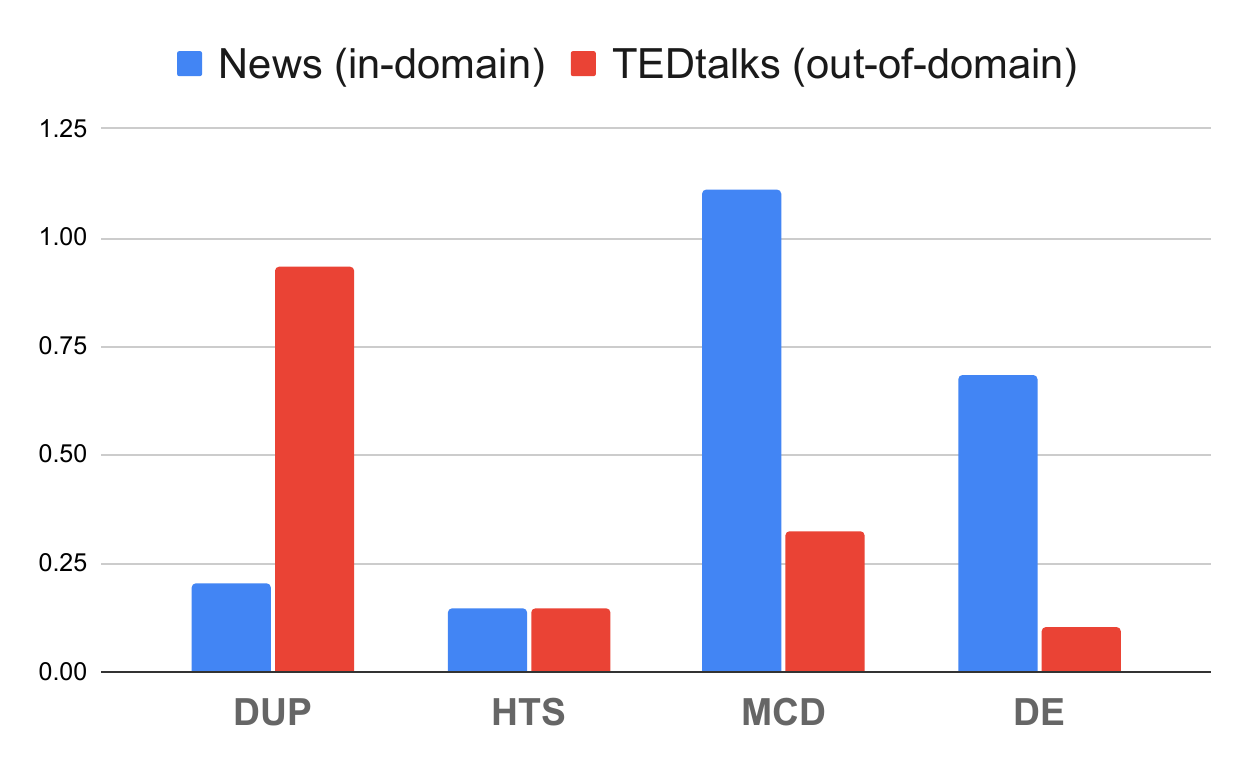}
\caption{\textbf{Epistemic uncertainty caused by out-of-domain data.} We show sharpness (average uncertainty) on two English-Russian test sets from the WMT21 metrics task: an in-domain dataset (News) and an out-of-domain dataset (TED talks). Our proposed method that handles epistemic uncertainty (Direct Uncertainty Predictor -- DUP) exhibits higher uncertainty on the out-of-domain dataset, as expected. The heteroscedastic (HTS) predictor, which detects aleatoric, but not epistemic uncertainty, has similar uncertainty in both datasets, and the MC dropout (MCD) and deep ensemble (DE) baselines,  surprisingly, has the opposite behavior.% \andre{maybe add News (in-domain) and TEDTalks (out-of-domain) to the the plot legend to be even clearer?}
} 
\label{fig:epistemic}
\end{figure}

\begin{figure*}[t]
\centering
\includegraphics[width=0.9\textwidth] {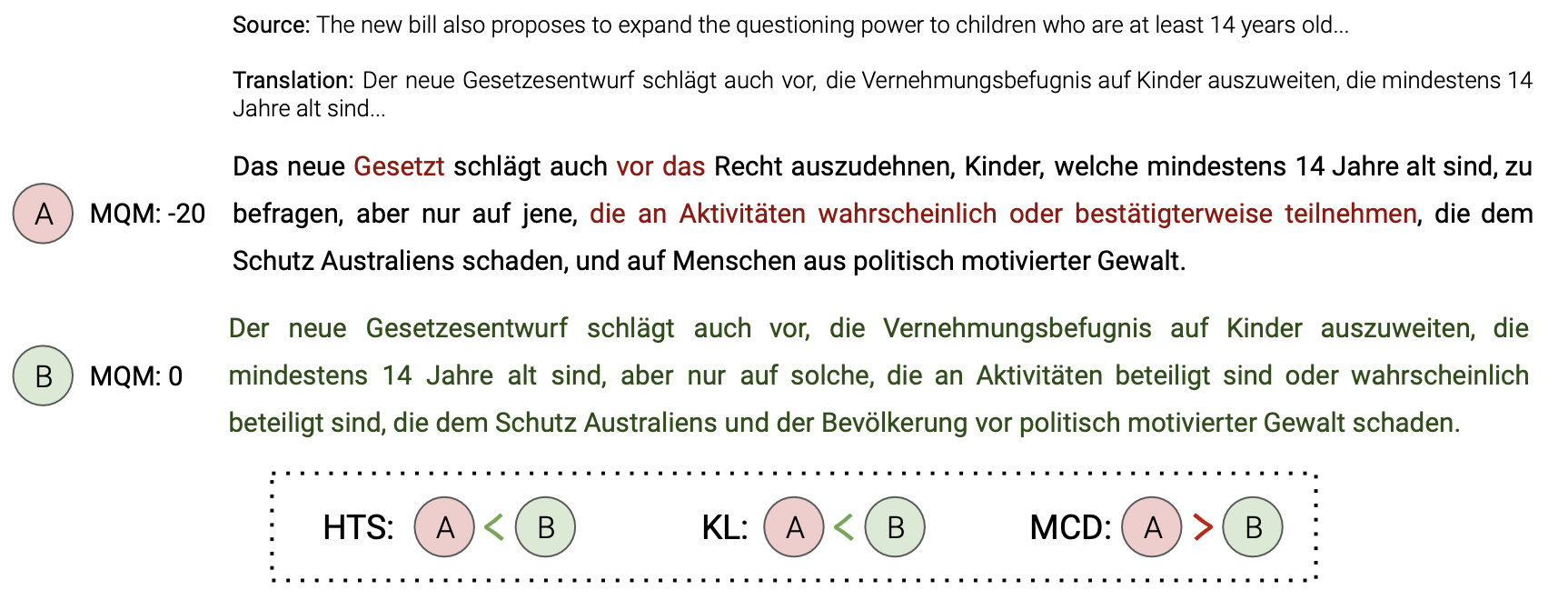}
\caption{\textbf{Aleatoric uncertainty caused by noisy references.} We show a low quality reference (A) and a high quality reference (B) for an English-German translation (the translation in the example is high quality, identical to reference B). Errors in reference A are annotated in \textcolor{red!70!black}{dark red};  
reference B has a perfect MQM score of 0 (no errors). 
Our two proposed methods that handle aleatoric (data) uncertainty, HTS and KL, are more uncertain when given the low-quality reference, as expected. The previously proposed MCD method \citep{glushkova-etal-2021-uncertainty} behaves in the opposite way. Full dataset statistics are shown in Figure~\ref{fig:ref2}.} 
\label{fig:ref1}
\end{figure*}

Trainable neural-based metrics, such as \textsc{Comet} or \textsc{Bleurt}~\citep{rei-etal-2020-comet, sellam-etal-2020-bleurt}, hold great promise for MT evaluation~\cite{freitag2021results}.  
For system comparison, 
they surpass or complement  
%They tend to complement and even replace 
traditional
lexical metrics such as \textsc{Bleu}~\cite{papineni-etal-2002-bleu}, and at a segment level, %these approaches, %they are showing great potencial 
%MT evaluation methods 
%based on large pretrained language models fine-tuned on human quality annotations, % have also shown great potential at a segment level, both 
they show higher correlations with human judgments, with and without access to references \cite{kepler-etal-2019-openkiwi, thompson-post-2020-automatic, ranasinghe-etal-2020-transquest}.

%\andre{when talking about these weaknesses, we could cite Chantal's paper} \andre{somewhere when talking about efficiency we could cite the Cometinho paper (or we could leave a note to cite it in the camera ready, to avoid too many self-citations)}
However, trainable MT evaluation metrics are not always trustworthy: For example, they can be unreliable in out-of-domain data and low resource languages, and sometimes they disregard specific error types, attributing high scores to low quality translations \cite{amrhein2022identifying}. Hence, we need a measure of \textbf{confidence} over their quality predictions for each segment, so that they can be better contextualized and interpreted. %Indeed, neural-based MT evaluation models are prone to multiple sources of epistemic (model) and aleatoric (data) uncertainty, often over- or under-estimating MT quality, especially when applied to new domains or languages.
Recently, \citet{glushkova-etal-2021-uncertainty} proposed \textbf{uncertainty-aware MT evaluation} by combining \textsc{Comet} with two simple uncertainty quantification methods, both based on model variance, namely, Monte Carlo (MC) dropout~\cite{gal2016dropout} and deep ensembles~\cite{lakshminarayanan2017simple}. However, these two methods have two important shortcomings: 
\begin{itemize}
    \item They are costly in terms of inference time (MC dropout) and training time (deep ensembles). %\andre{I'm wondering if we could report runtimes in one of the experiments to make this point stand out. Would we have convincing numbers?}\taya{yes, we are reporting this in section 5.1 ``computational costs''}
    % \item They require calibration to provide meaningful confidence estimates (confidence estimates that reflect the distance of predicted quality score to human evaluations) \andre{don't our proposed solutions need calibration too? if so, maybe remove this point.}
    \item They are not able to detect or distinguish between different sources of uncertainty. For example, it is impossible to infer whether the predicted uncertainty stems from a noisy and ambiguous reference, an out-of-distribution example, or noisy annotations. More fundamentally, they are highly model-dependent and cannot distinguish between aleatoric (data) and epistemic (model) uncertainty,  as illustrated in Figures~\ref{fig:epistemic}--\ref{fig:ref1}. 
    %\andre{This seems to be the most important point that we are addressing. Perhaps we can explain better what kinds of uncertainty MCD and DE do not handle well, so that we can show later how our proposed methods overcome these limitations}
\end{itemize}

In this paper, we address the limitations above by 
investigating more powerful and efficient uncertainty quantification methods: \textbf{direct uncertainty prediction}~\citep{jain2021deup}, a two-step approach which uses supervision over the quality prediction errors; \textbf{heteroscedastic regression}, which estimates input-dependent aleatoric uncertainty and can be combined with MC dropout~\cite{kendall2017}; and \textbf{divergence minimization}, which can estimate uncertainty from annotator disagreements, when multiple annotations are available for the same example. We examine the degree to which these predictors can \edit{improve segment-level uncertainty-aware MT evaluation} and target phenomena related to specific types of uncertainty: (i) aleatoric uncertainty in the case of heteroscedastic regression and divergence minimization, and (ii) epistemic uncertainty in the case of direct uncertainty prediction.

We evaluate our newly proposed uncertainty estimators on 16 language pairs from the WMT20 and WMT21 metrics shared task, using two types of human annotations: direct assessments (DA) and multi-dimensional quality metric scores (MQM). 
The experiments show that our estimators compare favourably against model variance baselines (MC dropout and deep ensembles), while being considerably faster. 
We also show that we can address specific issues for MT evaluation, such as detecting potentially incorrect references and out-of-distribution examples in the data, by choosing the most suitable uncertainty predictor among our proposed methods.%
%\footnote{Code and data available at: \url{https://github.com/deep-spin/uncertainties_MT_eval}}

%and predict uncertainty of a given MT evaluation system using supervision over the quality prediction errors. To that end, we compare different uncertainty-oriented features and loss functions, attempting to establish their contribution towards predicting and distinguishing uncertainty. We specifically focus on two different sources of uncertainty:
%\begin{enumerate}
%    \item Uncertainty stemming from out-of-distribution examples and
%    \item Uncertainty related to ambiguous or low quality references
%\end{enumerate} %of said MT evaluation systems.

% \begin{figure*}
%     \centering
%     XXX
%     %\includegraphics{}
%     \caption{\andre{This figure/table should demonstrate aleatoric vs epistemic uncertainty in MT evaluation}}
%     \label{fig:uncertainty}
% \end{figure*}

% \begin{figure*}[t]
% \centering
% \includegraphics[width=0.9\textwidth] {images/reference_example2.png}
% \caption{Example of low quality reference (A) versus high quality reference (B) for a given $\langle s, t\rangle$ pair. For reference A, \color{red}\textbf{Major errors} \color{black}are annotated with red, bold text and \color{red}Minor errors \color{black}with normal red text. Reference B contains no errors -- hence MQM is 0.\taya{add example of epistemic uncertainty en-de or zh-en}}
% %\chryssa{I am a bit worried this takes too much space, move to appendix?} \ricardo{It takes a lot of space but its a nice example}}
% \label{fig:ref1}
% \end{figure*}

\section{Related Work}
\label{sec:rw}

\paragraph{MT evaluation} 
Traditional metrics for MT evaluation are based on lexical overlap, including \textsc{Bleu}~\citep{papineni-etal-2002-bleu}, \textsc{Meteor}~\citep{banerjee-lavie-meteor2009}, and \textsc{chrF}~\citep{popovic-2015-chrf}. More recent metrics leverage large pretrained models, either unsupervised, such as \textsc{BertScore}~\citep{zhang2019bertscore}, \textsc{YiSi}~\citep{lo-2019-yisi} and \textsc{Prism}~\cite{thompson-post-2020-automatic}, or fine-tuned on human annotations, such as \textsc{Comet}~\citep{rei-etal-2020-comet} and \textsc{Bleurt}~\citep{sellam-etal-2020-learning}. In recent studies it has become increasingly evident that supervised metrics exhibit higher correlations with human judgements~\cite{mathur-etal-2020-results, freitag-etal-2021-experts} and produce more reliable assessments of MT quality~\cite{kocmi2021ship}. Nonetheless, all these metrics output a single point estimate, with the exception of \textsc{UA-Comet}~\citep{glushkova-etal-2021-uncertainty}, which returns a confidence interval along with a quality estimate. Our work builds upon \textsc{UA-Comet} by proposing improved uncertainty quantification. %, but it is applicable to any other evaluation metrics.

\paragraph{Uncertainty quantification} 
The problem of over-confident incorrect predictions affects neural models across tasks, and thus there are 
several works applying uncertainty quantification techniques to address this. Model variance methods such as MC Dropout \cite{gal2016dropout} and deep ensembles \cite{lakshminarayanan2017simple} have been applied on a range of  tasks to estimate the total uncertainty of a model. 
However, these methods are computationally costly to train and apply. \citet{malinin2019ensemble} propose to address this shortcoming with ensemble distribution distillation via prior networks (addressing the inference cost of ensembling). They further investigate the adaptation of the aforementioned method to regression problems \cite{malinin2020regression}, proposing two methods to estimate Regression Prior Networks (RPN), which however require either access to out-of-distribution data or the distillation of an ensemble of regression models into an RPN. 

Recently, \citet{pmlr-v161-ulmer21a} have shown that variance-based uncertainty estimation methods, which employ ensembling or MC dropout, can be unstable when applied to out-of-distribution data and often fail to provide accurate uncertainty estimates. \citet{raghu2019direct}, \citet{hu2021learning}, and \citet{jain2021deup} corroborate these findings and propose to train a direct epistemic uncertainty predictor on the errors of the main model as a better method to estimate epistemic uncertainty. 
%\citet{hu2021learning} also use a two-step approach and learn to predict epistemic uncertainty separately from the main model, applying their method on MRI reconstruction. 
To the best of our knowledge, direct uncertainty prediction has not been examined on MT evaluation (or other NLP tasks). 
%\chryssa{Note: I want to add \url{https://arxiv.org/pdf/2110.06427.pdf} in the stochastic uncertainty quantification methods for completeness -- they introduce an ensemble-based latent variable model (CVAE) to model both aleatoric and epistemic uncertainty} 
Contrary to epistemic uncertainty, aleatoric (data) uncertainty corresponds to the irreducible amount of prediction error(s), which is due to the noise present in the observed data. 
\citet{kendall2017} propose the use of heteroscedastic variance in the loss function. %, with added variance-related weight decay, that results in smaller updates for noisy data. 
\citet{wang2019aleatoric} propose a test-time augmentation-based aleatoric uncertainty. They compare and combine it with epistemic uncertainty, and show that it provides more representative uncertainty estimates than dropout-based ones alone. 
Our paper takes inspiration from these techniques to estimate aleatoric noise in MT evaluation.

\paragraph{Annotator disagreement} %\andre{we should cite here some NLP related works on this topic, Barbara Plank has some work, there's also some MT-related work that we could mention} \url{https://bplank.github.io/papers/eacl2014-costsensitive.pdf} \url{https://aclanthology.org/2021.naacl-main.204.pdf} \cite{fornaciari-etal-2021-beyond}
Several approaches have been proposed to understand and model annotator bias \citep{cohn2013modelling,hovy2021importance} and to leverage annotator disagreement in NLP applications~\citep{sheng2008get,plank2014learning,plank-etal-2016-multilingual,jamison2015noise,pavlick2019inherent}. Recently, soft-label multi-task learning objectives for classification tasks have been proposed by \citet{fornaciari-etal-2021-beyond}. Our Kullback-Leibler (KL) divergence minimization objective may be regarded as an extension of this approach for regression tasks, replacing (softmax) categoricals by Gaussian distributions. 

%\andre{merge with this: \textsc{Bleu} or \textsc{chrF}, were unsupervised and relied on lexical overlap \cite{papineni-etal-2002-bleu, popovic-2015-chrf}, 
%a recent line of work has shown that trained MT evaluation systems, such as \textsc{Comet} \cite{rei-etal-2020-comet} or \textsc{Bleurt} \cite{sellam-etal-2020-bleurt}, exhibit higher correlations with human judgements~\cite{freitag-etal-2021-experts, kocmi2021ship}.}

%\andre{should we cite PRISM here as well?}

\paragraph{Uncertainty in NLP} 
There are several works applying uncertainty quantification techniques to NLP, most commonly for (structured) classification tasks. \citet{fomicheva-etal-2020-unsupervised} use MC dropout to model MT confidence, and 
\citet{malinin2020uncertainty} study structured uncertainty estimation in autoregressive tasks, including MT and speech recognition. 
\citet{ye-etal-2021-towards} model uncertainty in performance prediction of NLP systems. \citet{mielke-etal-2019-kind} apply heteroscedastic models to assess language difficulty, whereas \citet{friedl-etal-2021-uncertainty} estimate aleatoric uncertainty in scientific peer reviewing. Recently, \citet{wang2022uncertainty} focus on calibration of regression models and show that uncertainty can be useful for data augmentation. 
Our paper also focuses on a regression task although some of our techniques and findings can apply more broadly to these problems.
%\andre{we should cite some papers here by Dennis Ulmer; we could also move Daniel Beck's citation here, and perhaps cite his new work: \url{https://direct.mit.edu/tacl/article/doi/10.1162/tacl_a_00483/111592/Uncertainty-Estimation-and-Reduction-of-Pre}}

% \andre{Taya, Chryssa, can you say 1-2 sentences about the main papers we should cover here and add them to the bib? then we can start structuring things. On heteroscedastic we should cite \citet{kendall2017}, maybe mention also this paper which uses applied heteroscedastic models in NLP -- \url{https://arxiv.org/pdf/1906.04726.pdf}.}
% \andre{On DEUP we should cite \citet{jain2021deup} and maybe \url{https://arxiv.org/abs/2002.05582} and other references cited in the related work of the DEUP paper. We shoudl briefly mention other work such as our EMNLP paper and Daniel Beck's work on Gaussian Processes for QE.}

%\andre{may want to integrate ``references may be guilty'' and Ott et al. ``Analyzing'' and the new Google TACL paper and Marcin's paper ``deploy or not deploy''}

\section{Uncertainty in MT Evaluation}
\label{sec:mt_evaluation}

\subsection{MT evaluation}

Throughout, we denote by $s$ a sentence in a source language, by $t$ its translation into a target language, and by $\mathcal{R}$ a set of reference translations. 
A segment-level \textbf{MT evaluation system} $\mathcal{M}_\mathrm{Q}$ (also called a ``translation quality metric'') is a system that takes as input a triple $\langle s, t, \mathcal{R} \rangle$ and outputs a quality score $\hat{q} \in \mathbb{R}$, reflecting how accurate $t$ is as a translation of $s$. \footnote{We focus on reference-based MT evaluation.} 
%When $\mathcal{R} = \varnothing$, the metric $\mathcal{M}_\mathrm{Q}$ is called referenceless;  otherwise it is reference-based. 
%MT evaluation systems can be unsupervised, for example if they rely on lexical overlap \cite{papineni-etal-2002-bleu, popovic-2015-chrf}  %\andre{BLUE, ChrF papers} 
%or on large pretrained models to identify the similarity between $t$ and $\mathcal{R}$ \cite{zhang2019bertscore,thompson-post-2020-paraphrase}, 

Current state-of-the-art evaluation metrics, such as 
\textsc{Comet} \cite{rei-etal-2020-comet} or \textsc{Bleurt} \cite{sellam-etal-2020-bleurt}, 
are trained with supervision on corpora annotated with human judgments $q^* \in \mathbb{R}$, such as direct assessments (DA; \citealt{graham-etal-2013-continuous}) or scores from multi-dimensional quality metric annotations (MQM; \citealt{lommel2014multidimensional}). 
This supervision encourages their predicted quality scores $\hat{q}$ to approximate the human perceived quality $q^*$, in a way that generalizes to unseen data.  
%\andre{perhaps write here the expression for generalization error if we need to point at it later}

%The goal of this system is to predict a quality score $\hat{q} \in \mathbb{R}$. Automatic MT evaluation systems (metrics) can be both unsupervised and supervised. The former rely on large language models to identify the similarity between $t$ and $\mathcal{R}$ \cite{zhang2019bertscore,thompson-post-2020-paraphrase}, 

%Our goal is to quantify the uncertainty of a

%A typical MT evaluation task is defined as follows: a system of choice takes as input a triple of segments $\langle s, t, \mathcal{R} \rangle$, where $s$ is a source sentence, $t$ is a translation sentence, and $\mathcal{R}$ is a set of references. The goal of this system is to predict a quality score $\hat{q} \in \mathbb{R}$. Automatic MT evaluation systems (metrics) can be both unsupervised and supervised. The former rely on large language models to identify the similarity between $t$ and $\mathcal{R}$ \cite{zhang2019bertscore,thompson-post-2020-paraphrase}, while the latter are trained on large corpora with ground truth scores $q^*$ like direct assessment (DA; \citealt{graham-etal-2013-continuous}) or multi-dimensional quality metric (MQM; \citealt{lommel2014multidimensional}), provided by human annotators \cite{rei-etal-2020-comet,}.

\subsection{Sources of uncertainty}\label{sec:sources_uncertainty}

While neural-based MT evaluation systems are more accurate than traditional lexical-based metrics such as \textsc{Bleu}, they are less transparent and may produce unreliable scores for out-of-domain inputs or when references are noisy~\cite{rei-etal-2020-unbabels, freitag2021results}. 
Our goal is to mitigate this problem by quantifying the \textbf{uncertainty} associated with their predicted scores. 
This uncertainty can come from several sources:
\begin{itemize}
    \item \textbf{Aleatoric (data) uncertainty} is primarily caused by noise in the data. Frequent sources of noise include inaccurate or inconsistent ground truth quality scores $q^*$ (usually noticeable from low inter-annotator agreement scores) and noisy reference translations $\mathcal{R}$,  which can mislead the MT evaluation system  \citep{freitag-etal-2020-bleu}. \item \textbf{Epistemic (model) uncertainty} reflects lack of knowledge from the model itself. This may be caused by limited training data, out-of-distribution examples (e.g., new languages, new domains, or diverse scoring schemes), or by complex, highly non-literal,  translations which may trigger weak spots in the MT evaluation model. 
\end{itemize}

Recently, \citet{glushkova-etal-2021-uncertainty} proposed an \textbf{uncertainty-aware} evaluation metric (\textsc{UA-Comet}) by experimenting with two simple uncertainty quantification techniques, MC dropout~\cite{gal2016dropout} and deep ensembles~\cite{lakshminarayanan2017simple}. Both techniques compute estimates based on \textbf{model variance} -- they estimate uncertainty by running multiple versions of the system (either produced on-the-fly with stochastic dropout noise or by using separate models trained with different seeds), and then computing the mean $\hat{\mu}$ and variance $\hat{\sigma}^2$ of the predicted scores. 
When given a triple $\langle s, t, \mathcal{R} \rangle$ as input, 
instead of returning a point estimate $\hat{q}$, \textsc{UA-Comet} treats the quality score as a random variable $Q$, modeled as a  Gaussian distribution $p_Q(q) = \mathcal{N}(q; \hat{\mu}, \hat{\sigma}^2)$. After a calibration step, the variance parameter of the Gaussian $\hat{\sigma}^2$ is used as the uncertainty estimate.

\section{Improving Uncertainty-Aware MT Evaluation}\label{sec:uncertainty}

%While the approach above can partially handle data and model uncertainty and estimate model confidence, And yet, it is hard to distinguish which types of uncertainty do they really handle, and more importantly – how well they do so.
A limitation of  \textsc{UA-Comet}  %\citet{glushkova-etal-2021-uncertainty}'s approach 
is its reliance on model variance techniques that often produce poor estimates of uncertainty  
%\andre{I don't think this ref to Malinin and Gales is the right one in this context, I was thinking about a ICLR 2020 paper pointing out limitations of MCD but maybe it's not theirs. maybe it's this? \url{https://openreview.net/pdf?id=BJxI5gHKDr}; but this one seems to be more about pitfall in evaluation metrics for uncertainty quantification} 
and conflate aleatoric and epistemic uncertainty, making it hard to accurately represent uncertainty related to out-of-distribution samples \cite{jain2021deup,zhang2021dense}. 
We therefore examine alternate methods to learn aleatoric and epistemic uncertainty directly from the available data. We assume that for each of the training scenarios and learning objectives described in the following sections, we can learn to predict the uncertainty of quality estimates $\hat{q}$ either as the noise variance $\sigma$ in the case of aleatoric uncertainty, or as the generalization error $\epsilon$ in the case of epistemic (and total) uncertainty.
% Point out MCD and DEE as uncertainty quantification techniques introduced in prior work that output a distribution with mean $\hat{\mu}$ and variance $\hat{\sigma}^2$ instead of a point estimate $\hat{q}$. Point out the limitations of those techniques (e.g. they can't handle well some types of uncertainty)

\subsection{Predicting aleatoric uncertainty}
\label{sec:aleatoric}
% 3.3 - Predicting Aleatoric Uncertainty with Heteroscedastic Regression -- explain homoscedastic vs heteroscedastic and why the latter makes sense for aleatoric uncertainty in MT evaluation, explain the heteroscedastic loss and provide intuition (maybe write the KL loss first? are we using the KL loss in the paper?)

%\andre{We should briefly recap here the 4 different kinds of uncertainty that we described in the EMNLP paper and that prior work used simple strategies such as MC Dropout and DEE to estimate this uncertainty, then point out the limitation of that approach and use it to motivate heteroscedastic regression and DEUP}

%\andre{Rather than being a property of the model, aleatoric uncertainty is a property of the data distribution and thus it can be learned as a function of the data \cite{kendall2017}. It can be categorised into homoscedastic and heteroscedastic uncertainty, depending on whether or not we can assume constant variance for each data instance. In most ML problems we can assume that we are dealing with heteroscedastic data, however most ML and statistical models simplify this assumption and ignore heteroscedasticity.}

%As mentioned in \S\ref{sec:rw}, 
Rather than a property of the model, aleatoric uncertainty is a property of the data distribution and thus it can be learned as a function of the data \cite{kendall2017}. 
It corresponds to uncertainty induced due to noise and inconsistencies. In the case of MT evaluation, we identify low quality references and inconsistent human annotations as the main sources of aleatoric uncertainty. The uncertainty associated with each data instance can vary: references have shown to be of different quality levels \cite{freitag-etal-2020-bleu}, while the quality scores depend largely on the annotators who sometimes have high disagreement \cite{toral2020reassessing}. 

\paragraph{Heteroscedasticity} 
%A group of random variables is  called homoscedastic if they have the same variance; otherwise they are \textbf{heteroscedastic}. 

A common assumption in regression problems (of which MT evaluation is an example) is that the noise in the data has constant variance throughout the dataset -- i.e., that the data is \textit{homoscedastic}. The mean squared error loss, for example, corresponds to the maximum likelihood criterion under Gaussian noise with fixed variance. 
However, this is not a suitable assumption in several problems, including MT evaluation, where real data is often \textbf{heteroscedastic} -- for example,  complex sentences requiring specific background knowledge may be subject to larger annotation errors (higher disagreement among annotators) and higher chance for noisy references than simpler sentences.  Therefore, the aleatoric uncertainty should be larger in those cases. 
%To properly model this heteroscedascity in the data, we need heteroscedastic models, which we next describe. 

\paragraph{Heteroscedastic regression}
We model aleatoric uncertainty as observation noise by training a model to predict not only a quality score for each triple, but also a variance estimate $\hat{\sigma}^2$ for this score. Under our heteroscedastic assumption, we assume that the variance is specific to each data sample and can be learned as a function of the data. 
We follow \citet{quoc2005heteroscedastic} and \citet{kendall2017} and incorporate  $\hat{\sigma}^2$ as part of the training objective, while learning the MT evaluation model parameters. 

Formally, let $x := \langle s, t, \mathcal{R} \rangle$ denote an input triple, as described in \S\ref{sec:mt_evaluation}. Our heteroscedastic uncertainty-aware MT evaluation system $\mathcal{M}_\mathrm{Q}^\mathrm{HTS}$ is a neural network that takes $x$ as input and outputs a mean score $\hat{\mu}(x)$ and a variance score $\hat{\sigma}^2(x)$ -- in practice, this is done by taking a \textsc{Comet} model and changing the output layer to output two scores ($\hat{\mu}(x)$ and $\log \hat{\sigma}^2(x)$) instead of one ($\hat{q}(x)$). 
This predicted mean and variance parametrize a Gaussian distribution $\hat{p}_Q(q | x; \theta) = \mathcal{N}(q; \hat{\mu}(x; \theta), \hat{\sigma}^2(x; \theta))$, where $\theta$ are the model parameters. 
Given a training set $\mathcal{D} = \{(x_1, q_1^*), \ldots, (x_N, q_N^*)\}$, 
the maximum likelihood training criterion amounts to maximize 
\begin{align}
    %\lefteqn{\frac{1}{N}\sum_{i=1}^N\log p_Q({q_i^*} |  x_i; \theta)=} \nonumber\\ &= 
    \lefteqn{\frac{1}{N}\sum_{i=1}^N\log \underbrace{\mathcal{N}({q_i^*}; \hat{\mu}(x_i, \theta), \hat{\sigma}^2(x_i, \theta))}_{p_Q({q_i^*} |  x_i; \theta)}=}\\
    &= -\frac{1}{N}\sum_{i=1}^N \mathcal{L}_{\mathrm{HTS}}(\hat{\mu}(x_i, \theta), \hat{\sigma}^2(x_i, \theta); q_i^*) + \mathrm{const.},\nonumber
\end{align}
%$ \frac{1}{N}\sum_{i=1}^N\log p_Q({q_i^*} |  x_i; \theta) = \frac{1}{N}\sum_{i=1}^N\log \mathcal{N}({q_i^*}; \hat{\mu}_i, \hat{\sigma}_i^2)$, which is the same as minimizing 
%$\frac{1}{N}\sum_{i=1}^N \mathcal{L}_{\mathrm{HTS}}(\hat{\mu}_i, \hat{\sigma}_i^2; q_i^*)$, 
where $\mathcal{L}_{\mathrm{HTS}}$ denotes the \textbf{heteroscedastic loss}: %
%\footnote{In practice, for numerical stability, we make a substitution $s := \log \hat{\sigma}^2$ leading to the following expression for the loss:
%\begin{align}
%\label{eq:hts2}
%\mathcal{L}_{\mathrm{HTS}}(\hat{\mu}, s; q^*) &= 
%\frac{1}{2}\left( \exp(-s) (q^* - %\hat{\mu})^2  + s\right).
%\end{align}} %
%solve the following optimization problem:
%\begin{align}
%\label{eq:hts1}
%\hat{\theta}_{\mathrm{HTS}} &= \arg\max_\theta \frac{1}{N}\sum_{i=1}^N\log p_Q({q_i^*} \mid x_i; \theta)\nonumber\\
%&= \arg\max_\theta \frac{1}{N}\sum_{i=1}^N\log \mathcal{N}({q_i^*}; \hat{\mu}_i, \hat{\sigma}_i^2) \nonumber\\
%&= \arg\min_\theta \underbrace{\frac{1}{N}\sum_{i=1}^N \left( \frac{(q_i^* - \hat{\mu}_i)^2}{2\hat{\sigma}_i^2} + \frac{1}{2}\log \hat{\sigma}_i^2 \right)}_{\mathcal{L}_{\mathrm{HTS}}(\theta)}
%\end{align}
\begin{align}
\label{eq:hts1}
\mathcal{L}_{\mathrm{HTS}}(\hat{\mu}, \hat{\sigma}^2; q^*) &= \frac{(q^* - \hat{\mu})^2}{2\hat{\sigma}^2} + \frac{1}{2}\log \hat{\sigma}^2.
\end{align}
%where $\theta$ denotes the model parameters and we use the short-notation 
%$\hat{\mu}_i \equiv \hat{\mu}(x_i; \theta)$ and $\hat{\sigma}_i^2 \equiv \hat{\sigma}^2(x_i; \theta)$. These two quantities are output by a neural network with parameters $\theta$ and input $x_i$. 
We can see that, if $\hat{\sigma}^2$ was constant and not estimated, the heteroscedastic loss $\mathcal{L}_\mathrm{HTS}$ would revert to a standard squared loss; however, since this variance is predicted by the model and changes with the input, the model is trained to make a trade-off: the $\hat{\sigma}^2$ term in the denominator down-weights examples where the target $q^*$ is assumed unreliable,  decreasing the impact of highly noisy instances (a form of weighted least squares), while the $\log \hat{\sigma}^2$ term penalizes the model if it overestimates the variance. 
We show in \S\ref{sec:noisy_refs} how this variance can be used to detect possibly noisy references. % \andre{todo} 

%\begin{align}
%\label{eq:hts2}
%\mathcal{L}_{\mathrm{HTS}}(\theta) &= 
%\frac{1}{2N}\sum_{i=1}^N \left( \exp(-s_i) (q_i^* - \hat{\mu}_i)^2  + s_i \right).
%\end{align}

%\begin{equation}
%\label{eq:hts1}
%    \mathcal{L}_\mathrm{HTS} = \frac{1}{N} \sum_{i=1}^N \frac{\|\vectorbold{q}^*_i-\vectorbold{\hat{q}}\|^2}{2 \vectorbold{\sigma}^2} +\frac{1}{2} \log \vectorbold{\sigma}^2
%\end{equation}

% \begin{equation}
% \label{eq:hts1}
%     \mathcal{L}_\mathrm{HTS} = \frac{1}{2N} \sum_{i=1}^N \frac{\|\vectorbold{y}_i-f(\vectorbold{x}_i)\|^2}{2 \vectorbold{\sigma} (\vectorbold{x}_i)^2} + \log \vectorbold{\sigma}(\vectorbold{x}_i)^2
% \end{equation}

%\begin{equation}
%\label{eq:hts2}
%    \mathcal{L}_\mathrm{HTS} = \frac{1}{N} \sum_{i=1}^N \frac{1}{2} \exp(-\vectorbold{s}_i) \|\vectorbold{q}^*_i-\vectorbold{\hat{q}}\| +\frac{1}{2} \vectorbold{s}_i
%\end{equation}

%\andre{Don't point to equations before we define them. 
%We also need to provide intuition about where \eqref{eq:hts1} comes from. Also, I am not sure we need to define heteroscedastic models in general before we apply them to MT evaluation. A more direct path is for \S 3 introduce the MT evaluation problem and notation (as we did in the EMNLP paper), and then stick to that notation when describing heteroscedastic models and DEUP.}

\paragraph{KL divergence minimization}
%
%Disentangling these two uncertainty sources can be quite complicated in absence of additional information. We thus consider two approaches towards predicting aleatoric uncertainty depending on the available information during training the MT evaluation model. In both cases described below we assume that we can learn to predict the data uncertainty during the training of the MT evaluation model. 
While heteroscedastic uncertainty allows to estimate the observation noise, when we have multiple annotations for the same example we may have additional information on data uncertainty reflected in \textbf{annotator disagreement}. We assume that annotator disagreement in this case can be used as a proxy to data uncertainty.  %that stems from noise and inconsistencies in the source and hypothesis segments. 

Similarly to the estimation of heteroscedastic variance with the $\mathcal{L}_\mathrm{HTS}$ objective, we assume that we can learn the variance  $\hat{\sigma}(x; \theta)$ as an estimator of aleatoric uncertainty alongside the rest of the model, but now leveraging the supervision coming from the annotator disagreement %\cite{zerva2021unbabel} % \andre{put this back in the final version}
-- we denote this system by $\mathcal{M}_\mathrm{Q}^\mathrm{KL}$. We %use the disagreement in annotator scores (computed as standard deviation values) as the aleatoric uncertainty target for each segment, and we 
model the annotator scores as another Gaussian distribution $p_Q^*(q \mid x) = \mathcal{N}(q; \mu^*(x), \sigma^*(x))$, where $\mu^*(x)$ is the sample mean and $\sigma^*(x)$ the sample variance of the annotator scores for the example $x$, used as targets for our model predictions. We formalize this as a KL divergence objective between the target distribution $p_Q^*$ and the predicted distribution $\hat{p}_Q$, which has the following closed form for Gaussian distributions:
\begin{align}
\label{eq:kl}
\lefteqn{\mathcal{L}_{\mathrm{KL}}(\hat{\mu}, \hat{\sigma}^2; \mu^*, {\sigma^*}^2) = \mathrm{KL}(p_Q^* \| \hat{p}_Q)} \nonumber\\
&= \frac{(\mu^* - \hat{\mu})^2 + {\sigma^*}^2}{2\hat{\sigma}^2} + \frac{1}{2}\log \frac{\hat{\sigma}^2}{{\sigma^*}^2} - \frac{1}{2}.
\end{align}
%\begin{align}
%\label{eq:kl}
%\mathcal{L}_\mathrm{KL}(\theta) =\nonumber\\ \frac{1}{N}\sum_{i=1}^N \left( \frac{(\mu_i^* - \hat{\mu}_i)^2 + {\sigma_i^*}^2}{2\hat{\sigma}_i^2} + \frac{1}{2}\log \frac{\hat{\sigma}_i^2}{{\sigma^*_i}^2} - \frac{1}{2}\right).
%\end{align} 
%\begin{equation}
%    \label{eq:kl}
%    \mathcal{L}_\mathrm{KL} = \log \frac{\hat{\sigma}}{\sigma^*} + \frac{{\sigma^*}^2+(\mu^* - \hat{\mu})^2}{2 \hat{\sigma}^2}-\frac{1}{2}
%\end{equation} 
Note that Eq. \ref{eq:kl} is a generalization of Eq. \ref{eq:hts1}: if we assume a fixed zero-limit variance ${\sigma^*}^2  \rightarrow 0$, we recover Eq. \ref{eq:hts1} up to a constant. % so that: \\$\mathcal{L}_\mathrm{KL} \propto
% \log \frac{\hat{\sigma}}{2} + \frac{(q^* - \hat{q})^2}{2 \hat{\sigma}^2}$.
%\andre{shouldn't it be $\mu_1-\mu_2$ and not $\mu_1+\mu_2$ in this formula? We could make a point here that with $\sigma_1$ fixed and in the zero limit, this loss becomes equivalent to the heteroscedastic loss -- the scenario where we don't observe the variance in the annotations. By the way, are we using this KL loss anywhere in this paper? This would be applicable if we used the QE data as we have multiple annotators there.} \chryssa{we use it onlt when we train on MQM 2020 (currently table 4)}

\subsection{Predicting epistemic uncertainty}\label{sec:epistemic}

Epistemic (model) uncertainty can be observed mainly on out-of-sample and out-of-distribution instances, and manifests as the \textit{reducible} generalization error of the model -- in the presence of infinite training data and suitable model and learning algorithm, epistemic uncertainty could be reduced to zero \cite{postels2021practicality,jain2021deup}. 
We outline two procedures to estimate epistemic and total uncertainty, one combining MC dropout with the heteroscedastic loss \citep{kendall2017}, and another which estimates uncertainty directly as the generalization error \citep{jain2021deup}.

%\subsubsection{Estimation of epistemic uncertainty}
%\label{sec:ep_unc_baseline}

\paragraph{Heteroscedastic MC dropout} 
\label{sec:hts_mcd}
Given a way to estimate aleatoric uncertainty $\hat{\sigma}$, e.g., using Eqs.~\ref{eq:hts1} or \ref{eq:kl}, we can combine it with an estimator of epistemic uncertainty to obtain the total uncertainty over a sample. Assuming we have access to an MT evaluation model that is able to predict both a quality score $\hat{q}$ and an aleatoric uncertainty estimate $\hat{\sigma}$ -- such as the system $\mathcal{M}_\mathrm{Q}^\mathrm{HTS}$ described in \S\ref{sec:aleatoric} -- we can use a stochastic strategy such as MC dropout or deep ensembles to obtain a set $\mathcal{Q} = \{\hat{q}_1, \ldots, \hat{q}_M\}$ of quality estimates and $\Sigma = \{{\hat{\sigma}_1}^2, \ldots, {\hat{\sigma}_M}^2\} $ of variance estimates. 
Assuming $\mathcal{Q}$ is a sample drawn from a Gaussian distribution, %$\hat{p}_Q(q) = \mathcal{N}(q; \hat{\mu}, \hat{\sigma}^2)$, it allows us to estimate the parameters  ${\mu(\hat{q})}$ and ${\sigma(\hat{q})}^2$ as the sample mean and variance, respectively, with ${\sigma(\hat{q})}^2$ corresponding to epistemic uncertainty. 
the sample variance can be used as an estimator of epistemic uncertainty, and the sample mean of $\Sigma$ can be used as an estimator of aleatoric uncertainty \cite{kendall2017}.
We can then estimate the total uncertainty over the $M$ samples as the sum of epistemic and aleatoric uncertainties: 
%Specifically, we experiment with Monte Carlo Dropout (MCD) \cite{gal2016dropout} or deep ensembles (DE) \cite{lakshminarayanan2016simple} as uncertainty estimators, since they have already been used successfully to estimate uncertainty on MT evaluation \cite{glushkova-etal-2021-uncertainty-aware}. In this paper we use these two aforementioned approaches as our baselines.
%\chryssa{Do we need to explain this further and provide the derivation?}\taya{I think we definitely should put it in appendix}
% \begin{equation}
%     \label{eq:tot1}
%     \mathcal{U}_\mathrm{total} = \frac{1}{N} \sum_{i=1}^N {\hat{q}}_i^2 -(\frac{1}{N} \sum_{i=1}^N  {\hat{q}}_i)^2 + \frac{1}{N} \sum_{i=1}^N  {\hat{\sigma}}_i^2
% \end{equation}
%\taya{should we add some braces to define specific parts? example:}
\begin{align}
    \label{eq:tot1}
    \hat{\mathcal{U}}_\mathrm{total} &=
    \mathrm{Var}[\mathcal{Q}] + \mathbb{E}[\Sigma]
    \\
    &= \underbrace{\frac{1}{M} \sum_{j=1}^M {\hat{q}}_j^2 -\left(\frac{1}{M} \sum_{j=1}^M  {\hat{q}}_j\right)^2 }_{\text{epistemic}}+ \underbrace{\frac{1}{M} \sum_{j=1}^M {\hat{\sigma}}_j^2}_{\text{aleatoric}}.\nonumber
\end{align}
For the experiments presented in \S\ref{sec:experiments} we use this strategy with MC dropout applied to a model trained with heteroscedastic regression. %and represent $\hat{\mathcal{U}}_\mathrm{total}$ as HTS+MCD. 

\paragraph{Direct uncertainty prediction} 
\label{sec:dup_description}
%Instead of attempting to approximate the uncertainty of a prediction using the variance estimation methods discussed in the previous sections, 
An alternative is to consider the total uncertainty $\hat{\mathcal{U}}_\mathrm{total}$ as an approximation of the \textbf{generalization error} of the MT evaluation model $\mathcal{M}_\mathrm{Q}$. % -- the error due to aleatoric uncertainty (the generalization error achievable by a Bayes optimal predictor) plus the error due to epistemic uncertainty.
In this case, assuming access to $\mathcal{M}_\mathrm{Q}$'s  predictions ${\hat{q}}$ and the ground truth quality scores $q^*$ on a new (unseen) set of samples, we could learn to predict the total uncertainty \textbf{directly} as the error $\epsilon$ between the model predictions ${\hat{q}}$ and the true scores ${q}^*$, using the strategy recently proposed by \citet{jain2021deup}. 

As opposed to the previously described uncertainty estimation approaches, direct uncertainty prediction (DUP) is a \textbf{two-step process}, as we need to first obtain the model $\mathcal{M}_\mathrm{Q}$ that generates the  predictions ${\hat{q}}$ that will allow us to estimate the target errors in a second stage. Hence, we need access to two distinct datasets on which two separate models have to be trained. We assume a dataset $\mathcal{D}_\mathrm{Q}$ where $\mathcal{M}_\mathrm{Q}$ is trained (we use the vanilla \textsc{Comet} system), and another, disjoint dataset $\mathcal{D}_\mathrm{E}$ where we train a second system $\mathcal{M}_\mathrm{E}$  to predict the uncertainty/error of $\mathcal{M}_\mathrm{Q}$'s predictions. %, thus learning uncertainty as a two-step process. %\andre{we do this for DEUP, but not for heteroscedastic regression, where we don't need a data split. Maybe create two subsections in this section, one about ``Direct Uncertainty Prediction'' and one about ``Heteroscedastic Regression for Uncertainty Prediction'', and describe the detais of each. Then, point out that they are actually close for one particular kind of DEUP loss}. 
For this purpose, we use $\mathcal{M}_\mathrm{Q}$ to annotate $\mathcal{D}_\mathrm{E}$ with quality estimates $\hat{q}$, and then we calculate the ground truth error $\epsilon^*$ as the distance to the human quality scores $q^*$ for each segment in $\mathcal{D}_\mathrm{E}$, 
%\begin{equation}
$\epsilon^* = |\hat{q}-q^*|$.
%   \label{eq:err}
%\end{equation} 
We use $\epsilon^*$  as the target to train $\mathcal{M}_\mathrm{E}$, given inputs $\langle s, t, \mathcal{R}, \hat{q} \rangle$. 
Letting $\hat{\epsilon}$ correspond to the  uncertainty predicted by $\mathcal{M}_\mathrm{E}$ on a given input, 
we  define  $\mathcal{L}^\mathrm{E}_\mathrm{HTS}$ function for  $\mathcal{M}_\mathrm{E}$: %\andre{one possible place to trim is to leave only one of these losses here}:

\begin{align}
   \mathcal{L}^\mathrm{E}_{\mathrm{HTS}}(\hat{\epsilon}; \epsilon^*) &= \frac{(\epsilon^*)^2}{2 \hat{\epsilon}^2} + \frac{1}{2}\log(\hat{\epsilon})^2.
   \label{eq:loss3}
\end{align} 

 $\mathcal{L}^\mathrm{E}_\mathrm{HTS}$ is inspired by the heteroscedastic loss of Eq.~\ref{eq:hts1}, where %similarly to the estimation of variance $\sigma^2$ 
the model is discouraged from predicting too high uncertainty values because of the term $\log(\hat{\epsilon})^2$, while it will still try to predict high $\hat{\epsilon}$ values for the samples where the MT quality score is not close to the human evaluation. Therefore, this choice is akin to a two-step approach to heteroscedastic regression: one step to train the ``mean'' predictor and another step for training the variance predictor given the mean predictions, where the two steps are performed on different partitions of the dataset, $\mathcal{D}_\mathrm{Q}$ and $\mathcal{D}_\mathrm{E}$. We show in Appendix \ref{sec:ablation} that  $\mathcal{L}^\mathrm{E}_\mathrm{HTS}$ outperforms other loss functions.

\section{Experiments}\label{sec:experiments}
The main focus of our experiments is to investigate how the uncertainty estimators we explore in this paper compare to each other and against proposed variance-based methods. Our comparisons address  the accuracy of uncertainty predictions on MT evaluation datasets (\S\ref{sec:exp_gen}) as well as more specific concerns such as the performance on out-of-domain data (\S\ref{sec:exp_gen}), the ability to detect low quality references (\S\ref{sec:noisy_refs}), and the computational costs (\S\ref{sec:cost}).

\subsection{Experimental Setup}
\label{sec:setup}

We follow \citet{glushkova-etal-2021-uncertainty} and use \textsc{Comet} (v1.0) %\footnote{We use v1.0 for all experiments.} 
as the underlying architecture for our MT evaluation models, trained on  the data from the WMT17-WMT19 metrics shared task \cite{freitag2021results}. 
%We use the MC Dropout (MCD) and deep ensemble (DE) approaches as  baselines to compare against the uncertainty prediction methods presented in \S\ref{sec:uncertainty}. 
We consider two types of human judgments: direct assessments (DA)  and multi-dimensional quality metric scores (MQM).

\paragraph{Experiments on DA scores}
% All single-step methods are trained on the data from the WMT17-WMT19 metrics shared task. We use a 20\% WMT20 newstest dataset  %\andre{I added this, can you confirm this is correct? and what do we use for validation?} 
% For all methods apart from DUP, we use 20\% / 80\% split of WMT20 newstest dataset for testing and calibration, respectively. For DUP, the use of this dataset is the following: we use 70\% for training the error predictor, 20\% for testing \andre{the same 20\% split as the other methods? this is still confusing}\taya{yes} and 10\% for calibration.
% \andre{can't we just say that all methods use 20\% of the WMT20 newstest dataset for testing; all methods except DUP use the remaining 80\% for calibration, whereas DUP uses 70\% for training the error predictor and 10\% for calibration. BTW: this looks a bit unfair, I thought DUP was using part of the training data to train the error predictor. It's a bit hard to justify why the other methods need 80\% for calibration and only 20\% for testing} \chryssa{They don't really need 80\% for testing, they get very similar results for just 10\% calibration like DUP. But we had discussed that it is unfair not to use the remaining 70\% for these models, and it should be used, even if it is only for calibration. In retrospect, I think we should have used it for fine-tuning and not for calibration.}

We evaluate our models using 5-fold cross validation on the WMT20 dataset.\footnote{We ensure that triplets from the same document appear all in a single fold so that all folds are disjoint. All folds are balanced with respect to the percentage of documents/source segments available for each language pair.} % 
All single-step models are trained on the data from the WMT17-19 metrics shared task using the development folds (80\%) for calibration. For DUP  models, WMT17-19 is used to train the first step model $\mathcal{M}_\mathrm{Q}$ and the development folds of WMT20 are used both for training the second step of the model $\mathcal{M}_\mathrm{E}$ and for calibration. %This way we make sure that all methods have access to the same data.
The data encompasses 16 language pairs (per-language results listed in Tables ~\ref{tab:da_notqe_enxx}--\ref{tab:da_notqe_xxen} in Appendix~\ref{sec:app_da_experiments}), which we aggregate into two groups, \textsc{En-Xx} (out-of-English) and \textsc{Xx-En} (into-English). We report the balanced average across all language pairs (\textsc{Avg}). 
%The \textbf{\textsc{En-Xx}} group contains \textsc{En-Cs}, \textsc{En-De}, \textsc{En-Ja}, \textsc{En-Pl}, \textsc{En-Ru}, \textsc{En-Ta}, \textsc{En-Zh}. The \textbf{\textsc{Xx-En}}: \textsc{Cs-En}, \textsc{De-En}, \textsc{Ja-En}, \textsc{Km-En}, \textsc{Pl-En}, \textsc{Ps-En}, \textsc{Ru-En}, \textsc{Ta-En}, \textsc{Zh-En}.

%\andre{we need to say here which/how many LPs we used and that we aggregate all experiments in En-XX and XX-En, and report the balanced average as Avg.}\chryssa{can we not just point to the appendix?}\ricardo{or as footnote? this last paragraph looks weird and a bit redundant... The source/target languages are the same right? so maybe we can just say from/into \{cs, de, ja, pl, ru, ta, zh?\}}\taya{sets from/into are different, I would rather point to Tables ~\ref{tab:da_notqe_enxx} and ~\ref{tab:da_notqe_xxen} in Appendix}

% Specifically for the direct uncertainty prediction experiments described in section \ref{sec:deup1}.

%These datasets are first combined into one, then split equally into two disjunct parts ($\mathcal{D}_{unc}$ and $\mathcal{D}_{eval}$). Resulting datasets are balanced across language pairs.

\paragraph{Experiments on MQM scores}
We fine-tune all models on the entire WMT20 MQM dataset, which consists of MQM annotations for English-German (\textsc{En-De}) and Chinese-English (\textsc{Zh-En}). %The KL model (Eq.~\ref{eq:kl}) is trained directly on the WMT20 MQM dataset, since the WMT data with direct assessments does not include information on annotator disagreement that is used as target for the KL model training. %\andre{do we use anything for validation? and for calibration? and for training the error predictor?}. 
For DUP, we finetune the $\mathcal{M}_\mathrm{E}$ model on WMT20. For testing and calibration we use WMT21 metrics shared task dataset, which contains MQM annotations for the same language pairs, but also with an addition of English-Russian (\textsc{En-Ru}). We evaluate using 5-fold cross validation on the WMT21 MQM data as well, where the development folds are used for calibrating models for each language pair. We also provide the performance on WMT21 without any finetuning on MQM scores in Appendix~\ref{sec:app_mqm_experiments}. %We provide results for both DA prediction models \andre{not clear what this means. I guess it's the same models used in the DA experiments without any fine-tuning but evaluated according to MQM scores?} (presented in Table ~\ref{tab:mqm} in App.~\ref{sec:app_mqm_experiments}) and their fine-tuned and tested on MQMs versions (see Table ~\ref{tab:mqm_ft}).

% \andre{this is a table, not the appendix section -- we should say ``Tab. XXX in App. YYY''}

\paragraph{Models} 
In the experiments that follow we use as baselines the two variance-based methods proposed by \citet{glushkova-etal-2021-uncertainty}: a MC dropout model with 100 dropout runs (\textbf{MCD}) and a deep ensemble of 5 independent models (\textbf{DE}), as well as the fixed-variance simple baseline they proposed: $\sigma_{\mathrm{fixed}}^2 = \frac{1}{|\mathcal{D}|} \sum_{{\langle s, t, \mathcal{R}, q^* \rangle \in \mathcal{D}}} (q^* - \hat{\mu})^2$. \\
We compare these baselines against our models:
\begin{itemize}
\setlength\itemsep{0em}
    \item \textbf{HTS}: The heteroscedastic model $\mathcal{M}_\mathrm{Q}^\mathrm{HTS}$ trained with the loss in Eq.~\ref{eq:hts1}.
    \item \textbf{HTS+MCD}: The combination of \textbf{HTS} with MC dropout as described 
in Eq.~\ref{eq:tot1}.
    \item \textbf{DUP}: The direct uncertainty prediction model described in \S\ref{sec:dup_description} using the $\mathcal{L}^\mathrm{E}_{\mathrm{HTS}}(\hat{\epsilon}; \epsilon^*)$ loss described in Eq. \ref{eq:loss3}.%\footnote{App. \ref{sec:ablation} shows that  $\mathcal{L}^\mathrm{E}_{\mathrm{HTS}}(\hat{\epsilon}; \epsilon^*)$  outperforms the models using other loss functions (Eqs. \ref{eq:loss1}, \ref{eq:loss2}). \andre{one possibility to save space is to only present this loss  and defer presentation and evaluation of the other losses to the appendix.}} %
    We use a vanilla \textsc{Comet} model as $\mathcal{M}_\mathrm{Q}$ and a system with the same architecture for $\mathcal{M}_\mathrm{E}$ which receives as an additional feature the predicted quality score $\hat{q}$ from $\mathcal{M}_\mathrm{Q}$. This extra feature is added by inserting a bottleneck layer between two feed-forward layers in the original \textsc{Comet} architecture (see App. ~\ref{sec:model_params}).
    \item \textbf{KL}: The divergence minimization model $\mathcal{M}_\mathrm{Q}^\mathrm{KL}$ using the objective in Eq.~\ref{eq:kl}. This model is used only for the experiments with MQM scores (Table \ref{tab:combined}), where multiple annotators for the same examples are available during training.%
    \footnote{Unlike the other models, the KL model is trained directly on the WMT20 MQM dataset (instead of being just fine-tuned there), since the WMT data with direct assessments does not include information on annotator disagreement that is used as target for the KL model training.}
\end{itemize}

\paragraph{Evaluation} To compare the performance of uncertainty predictors, we report the same performance indicators as \citet{glushkova-etal-2021-uncertainty}: the predictive Pearson score $r(\hat{\mu},q^*)$ (PPS), the uncertainty Pearson score $r(|q^*-\hat{\mu}|,\hat{\sigma})$ (UPS), the negative  log-likelihood $-\log \mathcal{N}(q^*; \hat{\mu}, \hat{\sigma}^2)$ (NLL), the expected calibration error (ECE; \citet{naeini2015obtaining}), and the sharpness (Sha.), i.e., the average predicted variance in the test set (see Appendix~\ref{sec:app_metrics} for details about these metrics). \edit{Note that we follow \cite{glushkova-etal-2021-uncertainty} in considering sharp confidence intervals desirable for all our in-domain experiments, however, for out-of-domain instances, the desired behaviour differs: we expect that the average predicted uncertainty on out-of-domain data would be higher compared to the average uncertainty observed on in-domain data. Hence higher sharpness values would be desirable in such cases (see also Figure \ref{fig:epistemic} and Appendix \ref{sec:app_ood_ext}).}

These indicators assess both quality prediction accuracy (PPS),  uncertainty-related accuracy (UPS) and calibration (ECE, Sha.), and the prediction and uncertainty accuracy combined in a single score (NLL). We consider UPS as our main indicator of performance, but report the other uncertainty indicators for completeness. PPS is reported as well, to assert that the performance of the quality predictions $\hat{q}$ of the MT evaluation model is not compromised. Additionally, we consider changes in average sharpness to be more indicative of the interpretability of the uncertainty predictions and the sensitivity of the model to domain and distribution shifts. We illustrate this  in Figure~\ref{fig:epistemic}.

% \subsection{Baselines}
% \begin{itemize}
%     \item MCD baseline \cite{}:
%     \begin{enumerate}
%         \item MCD for MT evaluation system trained over all available data and calibrated on val/test set with k-fold cross validation 
%         \item MCD for MT evaluation system that is trained on the train-1 part of the training data and calibrated on train-2
%     \end{enumerate}
%     \item Heteroschedastic loss \cite{kendall2017}:
%     \begin{enumerate}
%         \item trained over all available data and calibrated on val/test set with k-fold cross validation 
%         \item trained on the train-1 part of the training data and calibrated on train-2
%     \end{enumerate}
%     \item MCD + heteroschedastic loss:
%     Following \citep{kendall2017} we combine the variance predicted using the heteroschedastic loss objective with the MCD variance as follows \andre{what's inside the sum?}:
%     \begin{equation}
%         total\_uncertainty = \frac{1}{T} \sum_{t=1}^T 
%     \end{equation}
% \end{itemize}

%\section{Results and Discussion}

\subsection{Comparison of uncertainty methods}
\label{sec:exp_gen}
% \andre{we need to write some analysis of the results here refering to the two tables}

The results of the DA and MQM experiments are shown in Table~\ref{tab:combined}. 
As expected, the PPS values (which do not measure uncertainty, but accuracy of the quality predictions) are similar for all methods, since they are based either on a vanilla \textsc{Comet} model, or on an ensemble of \textsc{Comet} models, with an advantage for the \textbf{DE} method which benefits from the ensemble effect.  \textbf{HTS} and \textbf{KL}, which have modified objectives that learn the mean and the variance simultaneously, also boost PPS, but not as much as \textbf{DE}.
We focus our analysis on the uncertainty prediction accuracy, assessed primarily by UPS and also ECE, and Sharpness indicators.

\edit{For the DA experiments, we observe that our proposed methods, \textbf{HTS}, \textbf{DUP} and \textbf{KL}, show significantly \footnote{p<0.05 using William's test.} stronger uncertainty correlation (UPS) than the baseline estimates (\textbf{MCD} and \textbf{DE}), and obtain competitive scores for ECE, Sha. and NLL without significantly compromising PPS.}  % (see App.~\ref{sec:app_metrics}). 

Enhancing $\mathcal{M}_\mathrm{Q}^\mathrm{HTS}$ with MC dropout (\textbf{HTS+MCD}) seems to  further improve UPS and ECE, but produces less sharp uncertainty estimates and it negatively impacts the predictive accuracy. \textbf{DUP}'s main strength relates to provision of informative uncertainty intervals (changes in sharpness), while maintaining good performance for the other uncertainty metrics. As we can see in Figure \ref{fig:epistemic}, the sharpness of uncertainty predictions increases for out-of-domain data in the case of \textbf{DUP} and nicely captures the increased epistemic uncertainty in such cases. In contrast, we can see that variance-based epistemic uncertainty predictors are weaker in representing this domain shift and actually show the opposite behavior to the desired one, while aleatoric uncertainty (\textbf{HTS}) remains the same. We provide a more extended analysis of this aspect in  Appendix \ref{sec:app_ood_ext}. 
\edit{Additionally, we provide the performance of the uncertainty predictors when applied to two other metrics, \textsc{Bleurt} \cite{sellam-etal-2020-learning} and UniTE \cite{wan2022unite} in Appendix~\ref{app:sec:other_metrics}. We observe similar patterns, but notice that \textbf{HTS} approaches require access to source segments to provide meaningful uncertainty predictions.}

The findings on DA data are further supported by the MQM results especially for UPS, and we can see that the models achieve good performance for \textsc{En-Ru}, which is not available in the WMT20 MQM data used for fine-tuning (see Appendix \ref{sec:app_mqm_experiments}). We also see that the \textbf{KL} model, despite using significantly less training data (see \S\ref{sec:setup}), achieves competitive results and even outperforms other metrics for \textsc{En-De}.

\begin{table}[t]
\small
\centering
\addtolength{\tabcolsep}{-0.5pt}
\resizebox{7.7cm}{!}{
\begin{tabular}{clccccc}
\toprule
&  & UPS $\uparrow$ & ECE $\downarrow$ & Sha. $\downarrow$  & NLL $\downarrow$ & PPS $\uparrow$\\  

  %0.086 & 0.017 & \underline{0.524} & 1.473 & 0.287 \\
  \midrule
\multirow{6}{*}{\rotatebox{90}{\textsc{WMT20 DA}}}  
& $\sigma^2$-fixed  & --    & 0.019  & 0.415  & 1.236 & 0.444 \\
& MCD   & 0.106 & 0.016  & \underline{0.377}  & 1.199 & 0.443 \\
  &  DE  & 0.134 & 0.019  & 0.366  & \underline{1.156} & \underline{0.460} \\
  & HTS & 0.177 & 0.015  & 0.450  & 1.201 & 0.444 \\
  & HTS+MCD & \underline{0.254} & \underline{0.013} & 0.528 & 1.167  & 0.429 \\ 
  & DUP  & 0.182 &	0.014 & 0.437 &	1.190 &	0.444\\[0.1cm]
  %0.112 & 0.015 & \underline{0.411} & 1.266 & 0.446 \\ % 
  \toprule
  \midrule

\multirow{6}{*}{\rotatebox{90}{\textsc{WMT21 MQM}}}  
& $\sigma^2$-fixed  & --  & 0.055  & 0.371  & 2.090 & 0.377 \\
& MCD & 0.179 & \underline{0.024} & 0.334 & 1.686 & 0.460 \\
& DE & 0.128 & 0.051 & \underline{0.236} & 2.631 & \underline{0.479} \\
& HTS  & 0.307  & 0.041  & 0.284  & 2.264  & 0.445  \\
& HTS+MCD  & \underline{0.311}  & 0.037  & 0.388  & \underline{1.614}  & 0.445  \\
& KL  & 0.296  & 0.046  & 0.273  & 2.595  & 0.443  \\
& DUP  & 0.285  & 0.039  & 0.634  & 1.778  & 0.377  \\ %   HTS + DEUP + Features (Trained on WMT20*0.7)
  %& DUP_{f_\mathrm{\{Q,MCD\}}}^{L_3} &  &  &  &  &  \\
\bottomrule
\end{tabular}
}
\addtolength{\tabcolsep}{-0.5pt}
\caption{%\ricardo{Why is the PPS in the last column here and in the first column in Table 2} 
Results for segment-level DA and MQM predictions, averaged over all language pairs. \underline{Underlined} numbers indicate the best result for each evaluation metric in each language pair.}
\label{tab:combined}
\end{table}

\subsection{Identification of noisy references}
\label{sec:noisy_refs}

\begin{figure}[t]
\centering
\includegraphics[width=.90\columnwidth] {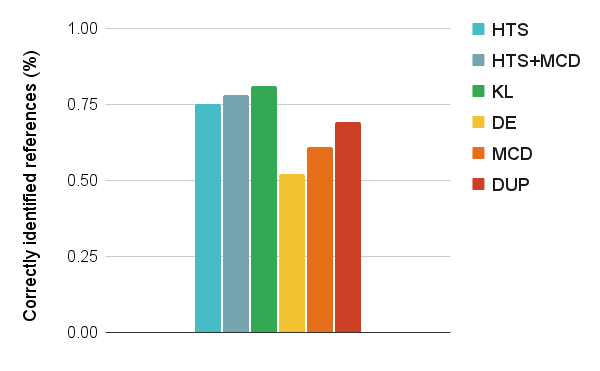}
\caption{Percentage of correctly recognized references with higher quality ($r_\mathrm{good}$ versus $r_\mathrm{bad}$) by different uncertainty predictors on the \textsc{En}-\textsc{De} dataset.}%\andre{do we have the DUP percentage somewhere?}}
\label{fig:ref2}
\end{figure}

As mentioned in \S\ref{sec:sources_uncertainty}, low quality references are a primary  source of aleatoric uncertainty. Thus, we  expect the uncertainty predictors that model aleatoric uncertainty (\textbf{HTS} and \textbf{KL}) to be more indicative of erroneous references compared to the other uncertainty predictors. % -- that is, we  expect them to predict higher uncertainty values when low-quality references are presented in the input. 
To verify this hypothesis, % and investigate the potential of aleatoric uncertainty predictors to detect noisy references, 
we conduct an experiment on the WMT21 MQM \textsc{En-De} dataset, which includes 4 references, each annotated with MQM scores by a human annotator \cite{freitag2021results}. %We can use these MQM scores as indicators of how good references are. 
For each $\langle s, t\rangle$ pair in the test split, we select the best reference $r_\mathrm{good}$ and the worst reference $r_\mathrm{bad}$ based on the respective MQM scores. 
%\andre{This triggered a question: For the previous DA and MQM experiments, do we use all references or just one of them?}\chryssa{Just one}
We retain only the $\langle s, t, \{r_\mathrm{good},r_\mathrm{bad}\} \rangle$ for which $|\mathrm{MQM}(r_\mathrm{good})-\mathrm{MQM}(r_\mathrm{bad})|>=10$, 
%\andre{we should mention above what is the scale of MQM scores -- 0--100? -- otherwise it's hard to understand this threshold}\chryssa{This is tricky because strictly speaking Google's MQM scores are not on a 0-100 scale. They just add penalty points according to identified errors. In practice it is between 0 and -25ish. I added a footnote, do you this it might be ok?}, 
so that there is a considerable quality difference between the references.% 
\footnote{An MQM penalty of 10 points corresponds to at least 2 major errors \citep{freitag-etal-2021-experts}.} %
We then apply the uncertainty predictors on the selected triples $\langle s, t, r_\mathrm{good}\rangle$ and $\langle s, t, r_\mathrm{bad}\rangle$ and obtain the predicted uncertainties, as shown in Figure \ref{fig:ref1}. For each $\langle s, t\rangle$ pair, we check which reference leads to the lowest predicted uncertainty and compute how often that reference coincides with $r_\mathrm{good}$. \edit{In Figure \ref{fig:ref2}, we can see that all the \textbf{HTS}, \textbf{HTS+MCD} and the \textbf{KL} predictors are much more successful in choosing the correct reference compared to \textbf{MCD}, \textbf{DE} and \textbf{DUP}.
This confirms the hypothesis that \textbf{HTS} and \textbf{KL} are more effective at capturing aleatoric uncertainty. Additionally, it is interesting to note that the combination of MC dropout with heteroscedastic loss provides a small boost to the accuracy of distinguishing the noisy reference.}

\subsection{Computational cost}
\label{sec:cost}

\begin{figure}[t]
\centering
\includegraphics[width=\columnwidth] {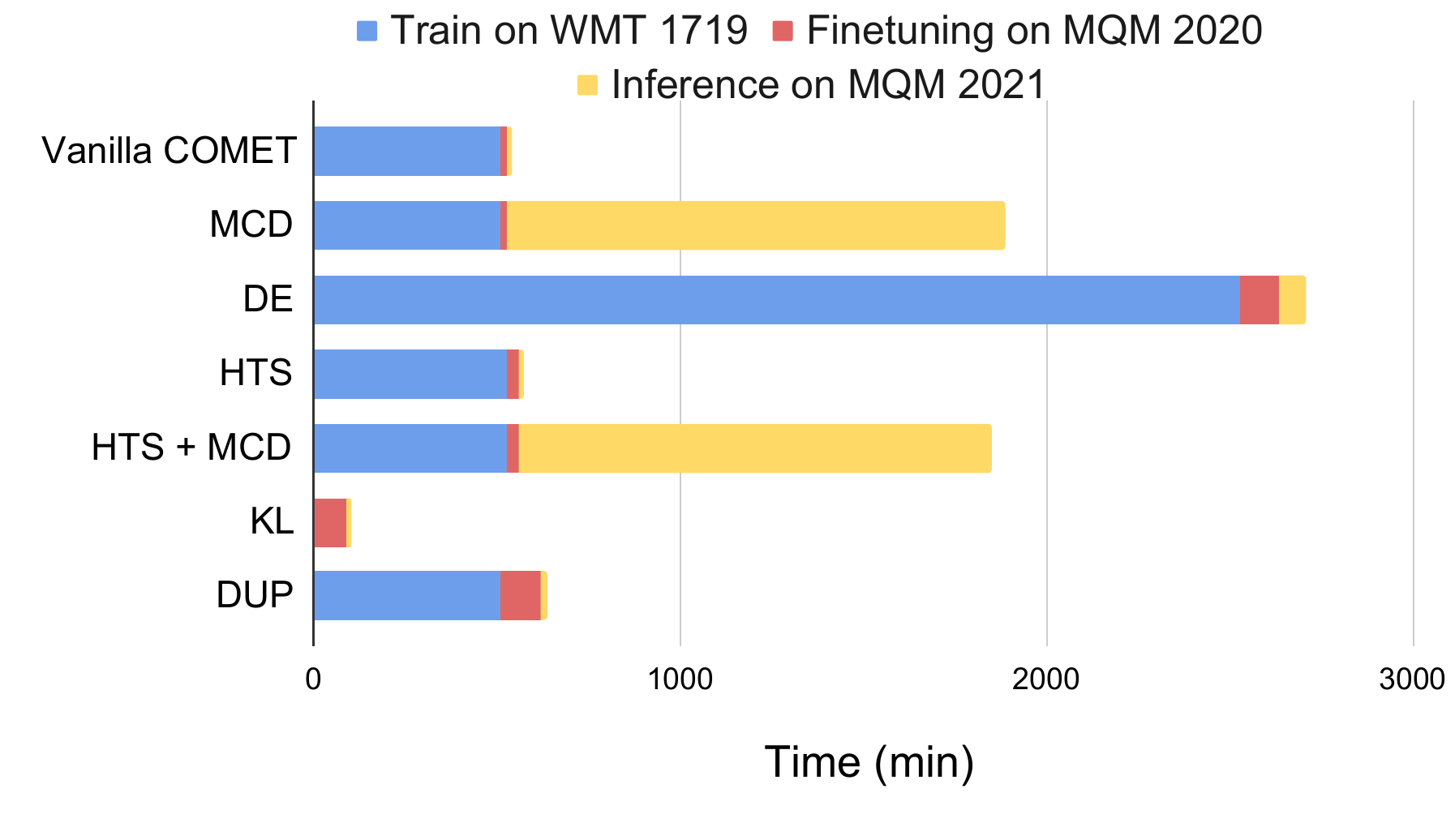}
\caption{Combined training, fine-tuning and inference times for the experiments reported in Table \ref{tab:combined}. All experiments were performed on a server with 4 Quadro RTX 6000 (24GB), 12 Intel Xeon Silver 4214@2.20GHz CPUs, and 256 Gb of RAM; time calculated for training/inference on a single GPU.}
\label{fig:time_plot}
\end{figure}

We now turn to the computational cost associated with the different uncertainty quantification methods, both in terms of training and inference runtime. 
%The number of models and the number of parameters for each model that are needed in order to acquire a quality score and an associated uncertainty estimate are an additional parameter to assess \andre{but we're not reporting the number of parameters, are we? the number of models we should report here -- i.e. how many models do we ensemble, how many MCD runs we need, etc. The MCD runs will look many, so we can point out that in general MCD requires many more runs than DE and point out to \citep[App. C]{glushkova-etal-2021-uncertainty} where it is shown that MCD degrades with fewer runs}, since recent MT evaluation models are particularly resource heavy. 
In Figure \ref{fig:time_plot}, we present the inference and training times for each of the  models (we used the same maximum number of epochs for each model).  %\andre{this is not necessarily fair unless it's a maximum number of epochs, since some models might require fewer epochs to learn}. 
The large inference times for  \textbf{MCD} and \textbf{HTS+MCD} stem from the need to perform 100 runs (the optimal number according to \citet{glushkova-etal-2021-uncertainty}); for DE, 5 models are ensembled, increasing training and inference costs 5-fold (for training details see Table~\ref{tab:hp} in Appendix~\ref{sec:model_params}). %The difference in number of quality estimates per sample between these two baselines might seem striking, but generally MCD requires many more runs than DE to achieve meaningful distributions.
In contrast, \textbf{HTS}, \textbf{KL}, and \textbf{DUP} have much lower costs (with slightly higher costs for \textbf{DUP} due to the need to train/run a second system) without performance compromises. 

%The fine-tuning time for \textbf{DUP} corresponds to the training of the $\mathcal{M}_\mathrm{E}$ model. This step requires the $\hat{q}$ estimates for each segment, i.e. one inference run with the vanilla COMET model over the dataset. This time has been aggregated with the total fine-tuning time for \textbf{DUP} \footnote{Similarly obtaining $\hat{q}$ estimates for the test set has been taken into account for the inference time calculated for DUP.}.  % \andre{why is this considered fine-tuning? it seems to me the cost of getting the $\hat{q}$ feature should be considered part of inference and not fine-tuning}.  
%However, this added inference time is practically negligible since it requires a single run per instance. \andre{we should add some more details here about the other methods, and why MCD and DE are so expensive. we should also explain the KL (or remove it). If we regenerate the figure, could we make the font bigger? Also, could we try to generate with a vector format (e.g. PDF) instead of PNG to avoid losing resolution?} 

\section{Conclusions}
We assessed the potential of different uncertainty predictors to capture different sources of uncertainty in MT evaluation.
We demonstrated that methods modeling heteroscedasticity can detect noisy references as a source of aleatoric uncertainty, and that the direct epistemic prediction method reflects well the increased epistemic uncertainty under a domain shift. 
Besides providing more informative uncertainty estimates than MC dropout and deep ensemble methods, our proposed predictors are also computationally cheaper. 

Overall, our work provides insight  about which uncertainty predictors  to choose for MT evaluation depending on the uncertainty source(s) to be addressed. \edit{The proposed uncertainty predictors that are able to target specific types of uncertainty are the first step towards mitigating the sources of such uncertainty, i.e. removing noisy instances from training to reduce aleatoric uncertainty, or identifying informative instances that would allow adapting to a new domain to reduce epistemic uncertainty. In future work, we are planning to further explore their properties and potential in improving MT and MT evaluation performance. }
%Thus, we argue that given a MT evaluation task and a neural metric, it is beneficial to choose one or more of the uncertainty predictors discussed in this work, depending on the uncertainty source(s) we want to address and the available resources.

%\andre{Thus with the training objectives in Eq. \ref{eq:loss1} to \ref{eq:loss3} we are learning to predict an upper bound of the total uncertainty. In an active learning setup where we can also measure the error after  $\mathcal{M}_\mathrm{Q}$ is trained on samples from  $\mathcal{D}_\mathrm{E}$ we could possibly get a better estimate of uncertainty \cite{jain2021deup}, but this scenario is out of the scope of this paper.}

\section*{Acknowledgements}

This work was supported by the European Research Council (ERC StG DeepSPIN 758969), by EU's Horizon Europe Research and Innovation Actions 
(UTTER, contract 101070631), by P2020 project MAIA (LISBOA-01-0247- FEDER045909), and Fundação para a Ciência e Tecnologia through contract UIDB/50008/2020.

% project PTDC/CCI-INF/4703/2021 (PRELUNA) and

%\clearpage
%\newpage

\section*{Limitations}

Our work addresses an important limitation of existing MT evaluation metrics -- the absence of reliable uncertainty estimates for their predictions and their inability to distinguish sources of uncertainty. 
However, our proposed approach has its own limitations as well. 
% \andre{This section does not count towards the page limit.}
First, the scope of our work is limited by the availability of resources with human quality annotations covering multiple languages. %, specially with respect to the languages considered for training and evaluation. 
Specifically, we are limited to the domains and language pairs addressed in the WMT metrics tasks (2017--2021) whose human assessments consist of DAs and MQM annotations. 
While these datasets include both high and low resource languages, most WMT datasets cover language pairs from or to English.
While certainly experimenting with more language pairs and domains in future work might provide additional insights, the WMT datasets used in our paper encompass 16 language pairs for testing and 24 for training, which still provides valuable information of variability across languages. %\andre{though they have a few non-EN LPs; did we use them in our experiments? it might look bad if we point this as a limitation and we didn't even use the ones that are available}\chryssa{They have only fr-de for 2020 but for some reason this data was not made available in official release} \andre{hmm, but IIRC the COMET paper reported results on non-English languages as well. Maybe we want to be a bit less explicit here not to raise red flags and just say that ``most WMT datasets cover language pairs from or to English''}. 
%Additionally, we only used the 2021 MQM annotations for testing, since the DA annotations were released later and based on the WMT Findings they correlate poorly with expert MQM annotations \cite{freitag2021results} \andre{is this relevant enough to point out?}. 
Second, the amount of sentences scored by more than one human annotator is scarce, and for this reason the experiments with the KL objective are  limited to a relatively small scale, which prevents a thorough comparison with the other uncertainty quantification methods. 
Third, while the uncertainty-related training objectives we propose are fully  general and can be applied to any supervised neural metric, we only experimented with \textsc{Comet} in this paper, due to limited computational resources. %\andre{due to limited computational resources? it may be good to make it sound like we had a good reason not to consider other metrics. ``Experimenting with other base metrics to see if they exhibit the same patterns is an interesting topic for future research.''}. 
Experimenting with other base metrics to see if they exhibit the same patterns is an interesting topic for future research. %As such, the discussion of results and conclusions relied on the impact of our proposed objectives on \textsc{Comet}. 
%Also, further work could be conducted to see if uncertainty predictions can help mitigate \textsc{COMET} weaknesses that were recently discussed in \citet{amrhein2022identifying} \andre{I think this is out of scope of this paper, we can probably omit this one. too many limitations might give the reviewers that our work could be improved with another round of revisions. the limitations we wanna point out here should be high level, and things for which we have a good reason not to have made. we should also point out at the end of each point something that we have done to mitigate these limitations. for example, when pointing out that we were limited to WMT datasets, we can say that ``While certaintly not covering many language pairs, the WMT datasets used in our paper encompass XXX language pairs, which still provides valuable information of variability across languages.''}. However we believe a proper analysis of these phenomena would merit a different paper, that also considers the word-level uncertainties and was thus out of scope for this work. 
Finally, our choice of uncertainty quantification techniques was guided by the desire to prioritize scalable and efficient methods that are applicable to different metrics and fit the MT evaluation task. 
Overall, we picked 6 different techniques (MCD, DE, HTS, HTS+MCD, KL, DUP) and left out other uncertainty quantification methods with less favorable efficiency or scalability properties.

%As such, we did not experiment with other methods for uncertainty quantification such as prior networks \cite{malinin2019ensemble}, variational inference and gaussian processes \cite{shen2019modelling, carvalho2020scalable} and Bayes-by-backprop \cite{ebrahimi2019uncertainty} methods. 
%Additionally, for the baselines we computed the variance of the distributions as the uncertainty value, because it aligned better with the definitions of uncertainty of our other predictors. While this is the most common approach \cite{fomicheva-etal-2020-unsupervised, gal2016dropout}, there is work that alongside the variance computes also the predictive entropy, and mutual information (MI) of the obtained MC Dropout samples \cite{nair2020exploring}. 

\bibliography{anthology,custom}

\begin{thebibliography}{51}
\expandafter\ifx\csname natexlab\endcsname\relax\def\natexlab#1{#1}\fi

\bibitem[{Amrhein and Sennrich(2022)}]{amrhein2022identifying}
Chantal Amrhein and Rico Sennrich. 2022.
\newblock Identifying {{Weaknesses}} in {{Machine Translation Metrics Through
  Minimum Bayes Risk Decoding}}: {{A Case Study}} for {{COMET}}.
\newblock In \emph{Proceedings of the 2nd Conference of the Asia-Pacific
  Chapter of the Association for Computational Linguistics and the 12th
  International Joint Conference on Natural Language Processing}, Online.
  Association for Computational Linguistics.

\bibitem[{Cohn and Specia(2013)}]{cohn2013modelling}
Trevor Cohn and Lucia Specia. 2013.
\newblock Modelling annotator bias with multi-task gaussian processes: An
  application to machine translation quality estimation.
\newblock In \emph{Proceedings of the 51st Annual Meeting of the Association
  for Computational Linguistics (Volume 1: Long Papers)}, pages 32--42.

\bibitem[{Fomicheva et~al.(2020)Fomicheva, Sun, Yankovskaya, Blain, Guzm{\'a}n,
  Fishel, Aletras, Chaudhary, and Specia}]{fomicheva-etal-2020-unsupervised}
Marina Fomicheva, Shuo Sun, Lisa Yankovskaya, Fr{\'e}d{\'e}ric Blain, Francisco
  Guzm{\'a}n, Mark Fishel, Nikolaos Aletras, Vishrav Chaudhary, and Lucia
  Specia. 2020.
\newblock \href {https://doi.org/10.1162/tacl_a_00330} {Unsupervised quality
  estimation for neural machine translation}.
\newblock \emph{Transactions of the Association for Computational Linguistics},
  8:539--555.

\bibitem[{Fornaciari et~al.(2021)Fornaciari, Uma, Paun, Plank, Hovy, and
  Poesio}]{fornaciari-etal-2021-beyond}
Tommaso Fornaciari, Alexandra Uma, Silviu Paun, Barbara Plank, Dirk Hovy, and
  Massimo Poesio. 2021.
\newblock \href {https://doi.org/10.18653/v1/2021.naacl-main.204} {Beyond black
  {\&} white: Leveraging annotator disagreement via soft-label multi-task
  learning}.
\newblock In \emph{Proceedings of the 2021 Conference of the North American
  Chapter of the Association for Computational Linguistics: Human Language
  Technologies}, pages 2591--2597, Online. Association for Computational
  Linguistics.

\bibitem[{Freitag et~al.(2021{\natexlab{a}})Freitag, Foster, Grangier,
  Ratnakar, Tan, and Macherey}]{freitag-etal-2021-experts}
Markus Freitag, George Foster, David Grangier, Viresh Ratnakar, Qijun Tan, and
  Wolfgang Macherey. 2021{\natexlab{a}}.
\newblock \href {https://doi.org/10.1162/tacl_a_00437} {Experts, errors, and
  context: A large-scale study of human evaluation for machine translation}.
\newblock \emph{Transactions of the Association for Computational Linguistics},
  9:1460--1474.

\bibitem[{Freitag et~al.(2020)Freitag, Grangier, and
  Caswell}]{freitag-etal-2020-bleu}
Markus Freitag, David Grangier, and Isaac Caswell. 2020.
\newblock \href {https://doi.org/10.18653/v1/2020.emnlp-main.5} {{BLEU} might
  be guilty but references are not innocent}.
\newblock In \emph{Proceedings of the 2020 Conference on Empirical Methods in
  Natural Language Processing (EMNLP)}, pages 61--71, Online. Association for
  Computational Linguistics.

\bibitem[{Freitag et~al.(2021{\natexlab{b}})Freitag, Rei, Mathur, Lo, Stewart,
  Foster, Lavie, and Bojar}]{freitag2021results}
Markus Freitag, Ricardo Rei, Nitika Mathur, Chi-kiu Lo, Craig Stewart, George
  Foster, Alon Lavie, and Ond{\v{r}}ej Bojar. 2021{\natexlab{b}}.
\newblock \href {https://aclanthology.org/2021.wmt-1.73} {Results of the
  {WMT}21 metrics shared task: Evaluating metrics with expert-based human
  evaluations on {TED} and news domain}.
\newblock In \emph{Proceedings of the Sixth Conference on Machine Translation},
  pages 733--774, Online. Association for Computational Linguistics.

\bibitem[{Friedl et~al.(2021)Friedl, Rizos, Stappen, Hasan, Specia, Hain, and
  Schuller}]{friedl-etal-2021-uncertainty}
Korbinian Friedl, Georgios Rizos, Lukas Stappen, Madina Hasan, Lucia Specia,
  Thomas Hain, and Bj{\"o}rn Schuller. 2021.
\newblock \href {https://doi.org/10.18653/v1/2021.findings-acl.443}
  {Uncertainty aware review hallucination for science article classification}.
\newblock In \emph{Findings of the Association for Computational Linguistics:
  ACL-IJCNLP 2021}, pages 5004--5009, Online. Association for Computational
  Linguistics.

\bibitem[{Gal and Ghahramani(2016)}]{gal2016dropout}
Yarin Gal and Zoubin Ghahramani. 2016.
\newblock Dropout as a bayesian approximation: Representing model uncertainty
  in deep learning.
\newblock In \emph{international conference on machine learning}, pages
  1050--1059. PMLR.

\bibitem[{Glushkova et~al.(2021)Glushkova, Zerva, Rei, and
  Martins}]{glushkova-etal-2021-uncertainty}
Taisiya Glushkova, Chrysoula Zerva, Ricardo Rei, and Andr{\'e} F.~T. Martins.
  2021.
\newblock \href {https://doi.org/10.18653/v1/2021.findings-emnlp.330}
  {Uncertainty-aware machine translation evaluation}.
\newblock In \emph{Findings of the Association for Computational Linguistics:
  EMNLP 2021}, pages 3920--3938, Punta Cana, Dominican Republic. Association
  for Computational Linguistics.

\bibitem[{Graham et~al.(2013)Graham, Baldwin, Moffat, and
  Zobel}]{graham-etal-2013-continuous}
Yvette Graham, Timothy Baldwin, Alistair Moffat, and Justin Zobel. 2013.
\newblock \href {https://aclanthology.org/W13-2305} {Continuous measurement
  scales in human evaluation of machine translation}.
\newblock In \emph{Proceedings of the 7th Linguistic Annotation Workshop and
  Interoperability with Discourse}, pages 33--41, Sofia, Bulgaria. Association
  for Computational Linguistics.

\bibitem[{Hovy and Yang(2021)}]{hovy2021importance}
Dirk Hovy and Diyi Yang. 2021.
\newblock The importance of modeling social factors of language: Theory and
  practice.
\newblock In \emph{Proceedings of the 2021 Conference of the North American
  Chapter of the Association for Computational Linguistics: Human Language
  Technologies}, pages 588--602.

\bibitem[{Hu et~al.(2021)Hu, Pezzotti, and Welling}]{hu2021learning}
Shi Hu, Nicola Pezzotti, and Max Welling. 2021.
\newblock Learning to predict error for mri reconstruction.
\newblock In \emph{International Conference on Medical Image Computing and
  Computer-Assisted Intervention}, pages 604--613. Springer.

\bibitem[{Jain et~al.(2021)Jain, Lahlou, Nekoei, Butoi, Bertin, Rector-Brooks,
  Korablyov, and Bengio}]{jain2021deup}
Moksh Jain, Salem Lahlou, Hadi Nekoei, Victor Butoi, Paul Bertin, Jarrid
  Rector-Brooks, Maksym Korablyov, and Yoshua Bengio. 2021.
\newblock Deup: Direct epistemic uncertainty prediction.
\newblock \emph{arXiv preprint arXiv:2102.08501}.

\bibitem[{Jamison and Gurevych(2015)}]{jamison2015noise}
Emily Jamison and Iryna Gurevych. 2015.
\newblock Noise or additional information? leveraging crowdsource annotation
  item agreement for natural language tasks.
\newblock In \emph{Proceedings of the 2015 Conference on Empirical Methods in
  Natural Language Processing}, pages 291--297.

\bibitem[{Kendall and Gal(2017)}]{kendall2017}
Alex Kendall and Yarin Gal. 2017.
\newblock What uncertainties do we need in bayesian deep learning for computer
  vision?
\newblock In \emph{NIPS}.

\bibitem[{Kepler et~al.(2019)Kepler, Tr{\'e}nous, Treviso, Vera, and
  Martins}]{kepler-etal-2019-openkiwi}
Fabio Kepler, Jonay Tr{\'e}nous, Marcos Treviso, Miguel Vera, and Andr{\'e}
  F.~T. Martins. 2019.
\newblock \href {https://doi.org/10.18653/v1/P19-3020} {{O}pen{K}iwi: An open
  source framework for quality estimation}.
\newblock In \emph{Proceedings of the 57th Annual Meeting of the Association
  for Computational Linguistics: System Demonstrations}, pages 117--122,
  Florence, Italy. Association for Computational Linguistics.

\bibitem[{Kocmi et~al.(2021)Kocmi, Federmann, Grundkiewicz, Junczys-Dowmunt,
  Matsushita, and Menezes}]{kocmi2021ship}
Tom Kocmi, Christian Federmann, Roman Grundkiewicz, Marcin Junczys-Dowmunt,
  Hitokazu Matsushita, and Arul Menezes. 2021.
\newblock \href {https://aclanthology.org/2021.wmt-1.57} {To ship or not to
  ship: An extensive evaluation of automatic metrics for machine translation}.
\newblock In \emph{Proceedings of the Sixth Conference on Machine Translation},
  pages 478--494, Online. Association for Computational Linguistics.

\bibitem[{Lakshminarayanan et~al.(2017)Lakshminarayanan, Pritzel, and
  Blundell}]{lakshminarayanan2017simple}
Balaji Lakshminarayanan, Alexander Pritzel, and Charles Blundell. 2017.
\newblock \href
  {https://proceedings.neurips.cc/paper/2017/file/9ef2ed4b7fd2c810847ffa5fa85bce38-Paper.pdf}
  {Simple and scalable predictive uncertainty estimation using deep ensembles}.
\newblock In \emph{Advances in Neural Information Processing Systems},
  volume~30. Curran Associates, Inc.

\bibitem[{Lavie and Denkowski(2009)}]{banerjee-lavie-meteor2009}
Alon Lavie and Michael Denkowski. 2009.
\newblock \href {https://doi.org/10.1007/s10590-009-9059-4} {The {M}eteor
  metric for automatic evaluation of {M}achine {T}ranslation}.
\newblock \emph{Machine Translation}, 23:105--115.

\bibitem[{Le et~al.(2005)Le, Smola, and Canu}]{quoc2005heteroscedastic}
Quoc~V. Le, Alex~J. Smola, and St\'{e}phane Canu. 2005.
\newblock \href {https://doi.org/10.1145/1102351.1102413} {Heteroscedastic
  gaussian process regression}.
\newblock In \emph{Proceedings of the 22nd International Conference on Machine
  Learning}, ICML '05, page 489–496, New York, NY, USA. Association for
  Computing Machinery.

\bibitem[{Lo(2019)}]{lo-2019-yisi}
Chi-kiu Lo. 2019.
\newblock \href {https://doi.org/10.18653/v1/W19-5358} {{Y}i{S}i - a unified
  semantic {MT} quality evaluation and estimation metric for languages with
  different levels of available resources}.
\newblock In \emph{Proceedings of the Fourth Conference on Machine Translation
  (Volume 2: Shared Task Papers, Day 1)}, pages 507--513, Florence, Italy.
  Association for Computational Linguistics.

\bibitem[{Lommel et~al.(2014)Lommel, Burchardt, and
  Uszkoreit}]{lommel2014multidimensional}
Arle Lommel, Aljoscha Burchardt, and Hans Uszkoreit. 2014.
\newblock \href {https://doi.org/10.5565/rev/tradumatica.77} {Multidimensional
  quality metrics ({MQM}): A framework for declaring and describing translation
  quality metrics}.
\newblock \emph{Tradumàtica: tecnologies de la traducció}, 0:455--463.

\bibitem[{Malinin et~al.(2020)Malinin, Chervontsev, Provilkov, and
  Gales}]{malinin2020regression}
Andrey Malinin, Sergey Chervontsev, Ivan Provilkov, and Mark Gales. 2020.
\newblock Regression prior networks.
\newblock \emph{arXiv preprint arXiv:2006.11590}.

\bibitem[{Malinin and Gales(2020)}]{malinin2020uncertainty}
Andrey Malinin and Mark Gales. 2020.
\newblock Uncertainty estimation in autoregressive structured prediction.
\newblock \emph{arXiv preprint arXiv:2002.07650}.

\bibitem[{Malinin et~al.(2019)Malinin, Mlodozeniec, and
  Gales}]{malinin2019ensemble}
Andrey Malinin, Bruno Mlodozeniec, and Mark Gales. 2019.
\newblock Ensemble distribution distillation.
\newblock In \emph{International Conference on Learning Representations}.

\bibitem[{Mathur et~al.(2020)Mathur, Wei, Freitag, Ma, and
  Bojar}]{mathur-etal-2020-results}
Nitika Mathur, Johnny Wei, Markus Freitag, Qingsong Ma, and Ond{\v{r}}ej Bojar.
  2020.
\newblock \href {https://aclanthology.org/2020.wmt-1.77} {Results of the
  {WMT}20 metrics shared task}.
\newblock In \emph{Proceedings of the Fifth Conference on Machine Translation},
  pages 688--725, Online. Association for Computational Linguistics.

\bibitem[{Mielke et~al.(2019)Mielke, Cotterell, Gorman, Roark, and
  Eisner}]{mielke-etal-2019-kind}
Sabrina~J. Mielke, Ryan Cotterell, Kyle Gorman, Brian Roark, and Jason Eisner.
  2019.
\newblock \href {https://doi.org/10.18653/v1/P19-1491} {What kind of language
  is hard to language-model?}
\newblock In \emph{Proceedings of the 57th Annual Meeting of the Association
  for Computational Linguistics}, pages 4975--4989, Florence, Italy.
  Association for Computational Linguistics.

\bibitem[{Naeini et~al.(2015)Naeini, Cooper, and
  Hauskrecht}]{naeini2015obtaining}
Mahdi~Pakdaman Naeini, Gregory~F. Cooper, and Milos Hauskrecht. 2015.
\newblock \href
  {https://people.cs.pitt.edu/~milos/research/AAAI_Calibration.pdf} {Obtaining
  well calibrated probabilities using bayesian binning}.
\newblock In \emph{Proceedings of the Twenty-Ninth AAAI Conference on
  Artificial Intelligence}, AAAI'15, page 2901–2907. AAAI Press.

\bibitem[{Papineni et~al.(2002)Papineni, Roukos, Ward, and
  Zhu}]{papineni-etal-2002-bleu}
Kishore Papineni, Salim Roukos, Todd Ward, and Wei-Jing Zhu. 2002.
\newblock \href {https://doi.org/10.3115/1073083.1073135} {{B}leu: a method for
  automatic evaluation of machine translation}.
\newblock In \emph{Proceedings of the 40th Annual Meeting of the Association
  for Computational Linguistics}, pages 311--318, Philadelphia, Pennsylvania,
  USA. Association for Computational Linguistics.

\bibitem[{Pavlick and Kwiatkowski(2019)}]{pavlick2019inherent}
Ellie Pavlick and Tom Kwiatkowski. 2019.
\newblock Inherent disagreements in human textual inferences.
\newblock \emph{Transactions of the Association for Computational Linguistics},
  7:677--694.

\bibitem[{Plank et~al.(2014)Plank, Hovy, and S{\o}gaard}]{plank2014learning}
Barbara Plank, Dirk Hovy, and Anders S{\o}gaard. 2014.
\newblock Learning part-of-speech taggers with inter-annotator agreement loss.
\newblock In \emph{Proceedings of the 14th Conference of the European Chapter
  of the Association for Computational Linguistics}, pages 742--751.

\bibitem[{Plank et~al.(2016)Plank, S{\o}gaard, and
  Goldberg}]{plank-etal-2016-multilingual}
Barbara Plank, Anders S{\o}gaard, and Yoav Goldberg. 2016.
\newblock \href {https://doi.org/10.18653/v1/P16-2067} {Multilingual
  part-of-speech tagging with bidirectional long short-term memory models and
  auxiliary loss}.
\newblock In \emph{Proceedings of the 54th Annual Meeting of the Association
  for Computational Linguistics (Volume 2: Short Papers)}, pages 412--418,
  Berlin, Germany. Association for Computational Linguistics.

\bibitem[{Popovi{\'c}(2015)}]{popovic-2015-chrf}
Maja Popovi{\'c}. 2015.
\newblock \href {https://doi.org/10.18653/v1/W15-3049} {chr{F}: character
  n-gram {F}-score for automatic {MT} evaluation}.
\newblock In \emph{Proceedings of the Tenth Workshop on Statistical Machine
  Translation}, pages 392--395, Lisbon, Portugal. Association for Computational
  Linguistics.

\bibitem[{Postels et~al.(2021)Postels, Segu, Sun, Van~Gool, Yu, and
  Tombari}]{postels2021practicality}
Janis Postels, Mattia Segu, Tao Sun, Luc Van~Gool, Fisher Yu, and Federico
  Tombari. 2021.
\newblock On the practicality of deterministic epistemic uncertainty.
\newblock \emph{arXiv preprint arXiv:2107.00649}.

\bibitem[{Raghu et~al.(2019)Raghu, Blumer, Sayres, Obermeyer, Kleinberg,
  Mullainathan, and Kleinberg}]{raghu2019direct}
Maithra Raghu, Katy Blumer, Rory Sayres, Ziad Obermeyer, Bobby Kleinberg,
  Sendhil Mullainathan, and Jon Kleinberg. 2019.
\newblock Direct uncertainty prediction for medical second opinions.
\newblock In \emph{International Conference on Machine Learning}, pages
  5281--5290. PMLR.

\bibitem[{Ranasinghe et~al.(2020)Ranasinghe, Orasan, and
  Mitkov}]{ranasinghe-etal-2020-transquest}
Tharindu Ranasinghe, Constantin Orasan, and Ruslan Mitkov. 2020.
\newblock \href {https://doi.org/10.18653/v1/2020.coling-main.445}
  {{T}rans{Q}uest: Translation quality estimation with cross-lingual
  transformers}.
\newblock In \emph{Proceedings of the 28th International Conference on
  Computational Linguistics}, pages 5070--5081, Barcelona, Spain (Online).
  International Committee on Computational Linguistics.

\bibitem[{Rei et~al.(2020{\natexlab{a}})Rei, Stewart, Farinha, and
  Lavie}]{rei-etal-2020-comet}
Ricardo Rei, Craig Stewart, Ana~C Farinha, and Alon Lavie. 2020{\natexlab{a}}.
\newblock \href {https://doi.org/10.18653/v1/2020.emnlp-main.213} {{COMET}: A
  neural framework for {MT} evaluation}.
\newblock In \emph{Proceedings of the 2020 Conference on Empirical Methods in
  Natural Language Processing (EMNLP)}, pages 2685--2702, Online. Association
  for Computational Linguistics.

\bibitem[{Rei et~al.(2020{\natexlab{b}})Rei, Stewart, Farinha, and
  Lavie}]{rei-etal-2020-unbabels}
Ricardo Rei, Craig Stewart, Ana~C Farinha, and Alon Lavie. 2020{\natexlab{b}}.
\newblock \href {https://aclanthology.org/2020.wmt-1.101} {Unbabel{'}s
  participation in the {WMT}20 metrics shared task}.
\newblock In \emph{Proceedings of the Fifth Conference on Machine Translation},
  pages 911--920, Online. Association for Computational Linguistics.

\bibitem[{Sellam et~al.(2020{\natexlab{a}})Sellam, Das, and
  Parikh}]{sellam-etal-2020-bleurt}
Thibault Sellam, Dipanjan Das, and Ankur Parikh. 2020{\natexlab{a}}.
\newblock \href {https://doi.org/10.18653/v1/2020.acl-main.704} {{BLEURT}:
  Learning robust metrics for text generation}.
\newblock In \emph{Proceedings of the 58th Annual Meeting of the Association
  for Computational Linguistics}, pages 7881--7892, Online. Association for
  Computational Linguistics.

\bibitem[{Sellam et~al.(2020{\natexlab{b}})Sellam, Pu, Chung, Gehrmann, Tan,
  Freitag, Das, and Parikh}]{sellam-etal-2020-learning}
Thibault Sellam, Amy Pu, Hyung~Won Chung, Sebastian Gehrmann, Qijun Tan, Markus
  Freitag, Dipanjan Das, and Ankur Parikh. 2020{\natexlab{b}}.
\newblock \href {https://aclanthology.org/2020.wmt-1.102} {Learning to evaluate
  translation beyond {E}nglish: {BLEURT} submissions to the {WMT} metrics 2020
  shared task}.
\newblock In \emph{Proceedings of the Fifth Conference on Machine Translation},
  pages 921--927, Online. Association for Computational Linguistics.

\bibitem[{Sheng et~al.(2008)Sheng, Provost, and Ipeirotis}]{sheng2008get}
Victor~S Sheng, Foster Provost, and Panagiotis~G Ipeirotis. 2008.
\newblock Get another label? improving data quality and data mining using
  multiple, noisy labelers.
\newblock In \emph{Proceedings of the 14th ACM SIGKDD international conference
  on Knowledge discovery and data mining}, pages 614--622.

\bibitem[{Thompson and Post(2020)}]{thompson-post-2020-automatic}
Brian Thompson and Matt Post. 2020.
\newblock \href {https://doi.org/10.18653/v1/2020.emnlp-main.8} {Automatic
  machine translation evaluation in many languages via zero-shot paraphrasing}.
\newblock In \emph{Proceedings of the 2020 Conference on Empirical Methods in
  Natural Language Processing (EMNLP)}, pages 90--121, Online. Association for
  Computational Linguistics.

\bibitem[{Toral(2020)}]{toral2020reassessing}
Antonio Toral. 2020.
\newblock \href {https://aclanthology.org/2020.eamt-1.20} {Reassessing claims
  of human parity and super-human performance in machine translation at {WMT}
  2019}.
\newblock In \emph{Proceedings of the 22nd Annual Conference of the European
  Association for Machine Translation}, pages 185--194, Lisboa, Portugal.
  European Association for Machine Translation.

\bibitem[{Ulmer and Cin\`a(2021)}]{pmlr-v161-ulmer21a}
Dennis Ulmer and Giovanni Cin\`a. 2021.
\newblock \href {https://proceedings.mlr.press/v161/ulmer21a.html} {Know your
  limits: Uncertainty estimation with relu classifiers fails at reliable ood
  detection}.
\newblock In \emph{Proceedings of the Thirty-Seventh Conference on Uncertainty
  in Artificial Intelligence}, volume 161 of \emph{Proceedings of Machine
  Learning Research}, pages 1766--1776. PMLR.

\bibitem[{Wan et~al.(2022)Wan, Liu, Yang, Zhang, Chen, Wong, and
  Chao}]{wan2022unite}
Yu~Wan, Dayiheng Liu, Baosong Yang, Haibo Zhang, Boxing Chen, Derek Wong, and
  Lidia Chao. 2022.
\newblock Unite: Unified translation evaluation.
\newblock In \emph{Proceedings of the 60th Annual Meeting of the Association
  for Computational Linguistics (Volume 1: Long Papers)}, pages 8117--8127.

\bibitem[{Wang et~al.(2019)Wang, Li, Aertsen, Deprest, Ourselin, and
  Vercauteren}]{wang2019aleatoric}
Guotai Wang, Wenqi Li, Michael Aertsen, Jan Deprest, S{\'e}bastien Ourselin,
  and Tom Vercauteren. 2019.
\newblock Aleatoric uncertainty estimation with test-time augmentation for
  medical image segmentation with convolutional neural networks.
\newblock \emph{Neurocomputing}, 338:34--45.

\bibitem[{Wang et~al.(2022)Wang, Beck, Baldwin, and
  Verspoor}]{wang2022uncertainty}
Yuxia Wang, Daniel Beck, Timothy Baldwin, and Karin Verspoor. 2022.
\newblock Uncertainty estimation and reduction of pre-trained models for text
  regression.
\newblock \emph{Transactions of the Association for Computational Linguistics},
  10:680--696.

\bibitem[{Ye et~al.(2021)Ye, Liu, Fu, and Neubig}]{ye-etal-2021-towards}
Zihuiwen Ye, Pengfei Liu, Jinlan Fu, and Graham Neubig. 2021.
\newblock \href {https://doi.org/10.18653/v1/2021.eacl-main.324} {Towards more
  fine-grained and reliable {NLP} performance prediction}.
\newblock In \emph{Proceedings of the 16th Conference of the European Chapter
  of the Association for Computational Linguistics: Main Volume}, pages
  3703--3714, Online. Association for Computational Linguistics.

\bibitem[{Zhang et~al.(2021)Zhang, Dai, Xiang, Fan, Moghadam, He, Walder,
  Zhang, Harandi, and Barnes}]{zhang2021dense}
Jing Zhang, Yuchao Dai, Mochu Xiang, Deng-Ping Fan, Peyman Moghadam, Mingyi He,
  Christian Walder, Kaihao Zhang, Mehrtash Harandi, and Nick Barnes. 2021.
\newblock Dense uncertainty estimation.
\newblock \emph{arXiv preprint arXiv:2110.06427}.

\bibitem[{Zhang et~al.(2019)Zhang, Kishore, Wu, Weinberger, and
  Artzi}]{zhang2019bertscore}
Tianyi Zhang, Varsha Kishore, Felix Wu, Kilian~Q Weinberger, and Yoav Artzi.
  2019.
\newblock Bertscore: Evaluating text generation with bert.
\newblock In \emph{International Conference on Learning Representations}.

\end{thebibliography}
\bibliographystyle{acl_natbib}

\clearpage
\newpage
\appendix
\section{DA experiments}
\label{sec:app_da_experiments}
Results per language pair are presented in Tables \ref{tab:da_notqe_enxx} and \ref{tab:da_notqe_xxen}.
%\chryssa{not done updating these}
\begin{table}[!h]
\small
\centering
\addtolength{\tabcolsep}{-0.5pt}
\resizebox{7.7cm}{!}{
\begin{tabular}{clccccc}
\toprule
&  & UPS $\uparrow$ & ECE $\downarrow$ & Sha. $\downarrow$   & NLL $\downarrow$ & PPS $\uparrow$ \\  
\midrule
\multirow{6}{*}{\rotatebox{90}{\textsc{En-Cs}}} 
& $\sigma^2$-fixed  & --    &  0.005  & 0.408  & 1.019 & 0.699 \\
& MCD  & 0.149 & 0.012  & 0.258  & 0.847 & 0.692   \\
  &  DE & 0.125 & 0.010  & \underline{0.255}  & \underline{0.825} & \underline{0.734} \\
  & HTS  & 0.105 & \underline{0.003}  & 0.456  & 1.025 & 0.699 \\
  & HTS+MCD & \underline{0.335} & 0.008 & 0.388 & 0.880 & 0.676 \\
  & DUP &  0.146 & \underline{0.003} & 0.419 & 1.010 & 0.699 \\
%   & DUP_{f_\mathrm{\{Q,MCD\}}}^{L_3} &  &  &  &  &   \\
\midrule
\multirow{6}{*}{\rotatebox{90}{\textsc{En-De}}}
& $\sigma^2$-fixed  & --    & 0.032  & \underline{0.172}  & 1.106 & 0.586 \\
& MCD  & 0.284 & 0.013  & 0.260  & \underline{0.916} & 0.582 \\
  &  DE & 0.209 & 0.025  & 0.198  & 0.949 & \underline{0.616} \\
  &  HTS  &  0.231 & 0.022  & 0.243  & 1.016 & 0.586 \\
  & HTS+MCD  & \underline{0.386} & \underline{0.004} & 0.369 & 0.924 & 0.612 \\ 
  & DUP & 0.232 & 0.022 & 0.190 & 1.028 & 0.586 \\
%   & DUP_{f_\mathrm{\{Q,MCD\}}}^{L_3}  &  &  &  &  & \\
\midrule
\multirow{6}{*}{\rotatebox{90}{\textsc{En-Ja}}} 
& $\sigma^2$-fixed  & --    & 0.011  & 0.186  & 0.714 & 0.636 \\
& MCD  & 0.236 & 0.005  & 0.225  & 0.697 & 0.638 \\
  &  DE & 0.140 & 0.010  & \underline{0.179}  & \underline{0.657} & 0.661 \\
  &  HTS  & 0.184 & 0.007  & 0.218  & 0.691 & 0.636\\
  & HTS+MCD   & \underline{0.249}  & \underline{0.003} & 0.520 & 1.048 & \underline{0.669} \\
  & DUP & 0.220 & 0.005 & 0.201 & 0.679 & 0.636 \\
%   & DUP_{f_\mathrm{\{Q,MCD\}}}^{L_3}  &  &  &  &  & \\
  \midrule
\multirow{6}{*}{\rotatebox{90}{\textsc{En-Pl}}}  
& $\sigma^2$-fixed  & --    & 0.010  & 0.410  & 1.090 & 0.609 \\
  & MCD & 0.167 & 0.010  & 0.326  & \underline{0.947} & 0.610  \\
  & DE & 0.129 & 0.011  & \underline{0.315}  & 0.960 & \underline{0.645} \\
  & HTS & 0.103 & 0.008  & 0.436  & 1.095 & 0.609 \\
  & HTS+MCD & \underline{0.189} & \underline{0.004} & 0.397 & 0.969 & 0.631\\
  & DUP  & 0.112 & 0.004 & 0.431 & 1.094 & 0.609 \\
%   & DUP_{f_\mathrm{\{Q,MCD\}}}^{L_3}  &  &  &  &  & \\
\midrule
\multirow{6}{*}{\rotatebox{90}{\textsc{En-Ru}}}
& $\sigma^2$-fixed  & --    & 0.018  & 0.287  & 0.989 & 0.534 \\
& MCD   & 0.200 & 0.012  & 0.307  & 0.943 & 0.538 \\
  &  DE & 0.146 & 0.016  & \underline{0.263}  & \underline{0.898} & \underline{0.570} \\
  &  HTS  & 0.156 & 0.012  & 0.321  & 1.006 & 0.534 \\
  & HTS+MCD  & \underline{0.370} & 0.017 & 0.637 & 1.050 & 0.497 \\
  & DUP & 0.154 & \underline{0.009} & 0.305 & 0.991 & 0.534 \\
%   & DUP_{f_\mathrm{\{Q,MCD\}}}^{L_3}  &  &  &  &  & \\
\midrule
\multirow{6}{*}{\rotatebox{90}{\textsc{En-Ta}}}
& $\sigma^2$-fixed  & --    &  0.017  & \underline{0.356}  & 1.077 & 0.661\\
& MCD   & 0.077 & \underline{0.004}  & 0.406  & 1.035 & 0.658  \\
  &  DE & 0.111 & 0.010  & 0.361  & \underline{1.030} & \underline{0.676} \\
  &  HTS  & 0.250 & 0.009  & 0.435  & 1.050 & 0.661\\
  & HTS+MCD  & 0.231 & 0.038 & 0.921 & 1.223 & 0.617 \\
  & DUP & \underline{0.252} & 0.009 & 0.385 & 1.032 & 0.661 \\
%   & DUP_{f_\mathrm{\{Q,MCD\}}}^{L_3}  &  &  &  &  & \\
\midrule
\multirow{6}{*}{\rotatebox{90}{\textsc{En-Zh}}}
& $\sigma^2$-fixed  & --    & 0.017  & 0.173  & 1.016 & 0.325 \\
& MCD  & 0.083 & 0.018  & \underline{0.152}  & 1.367 & 0.330  \\
  &  DE & 0.560 & 0.021  & \underline{0.152}  & 0.813 & 0.327 \\
  &  HTS  & 0.728 & \underline{0.003}  & 0.273  & 0.657 & 0.325 \\
  & HTS+MCD  & 0.504 & 0.020 & 0.380 & 0.865 & \underline{0.562} \\
  & DUP & \underline{0.722} & \underline{0.003} & 0.268 & \underline{0.650} & 0.325 \\ \midrule
%   & DUP_{f_\mathrm{\{Q,MCD\}}}^{L_3}  &  &  &  &  & \\ 
\multirow{6}{*}{\rotatebox{90}{\textsc{En-Xx}}} 
& $\sigma^2$-fixed  & --    & 0.015  & 0.288  & 1.000 & 0.566 \\
  & MCD   & 0.163 & 0.011  & 0.265  & 0.984 & 0.566  \\
  &  DE  & 0.223 & 0.015  & \underline{0.240}  & \underline{0.864} & 0.591\\
  & HTS   & 0.272 & 0.008  & 0.344  & 0.919 & 0.566\\
  & HTS+MCD  &	\underline{0.323} &	0.017 &	0.516 &	0.994 & \underline{0.609}\\
  & DUP   & 0.285 &	\underline{0.007} &	0.320 &	0.910 &	0.566\\
\bottomrule
\end{tabular}
}
\addtolength{\tabcolsep}{-0.5pt}
\caption{Results for segment-level DA prediction for En-Xx LPs. \underline{Underlined} numbers indicate the best result for each evaluation metric in each language pair.}
\label{tab:da_notqe_enxx}
\end{table}

\begin{table}[ht!]
\small
\centering
\addtolength{\tabcolsep}{-0.5pt}
\resizebox{7.7cm}{!}{
\begin{tabular}{clccccc}
\toprule
&   & UPS $\uparrow$   & ECE $\downarrow$ & Sha. $\downarrow$ & NLL $\downarrow$ & PPS $\uparrow$\\  
\midrule
\multirow{6}{*}{\rotatebox{90}{\textsc{Cs-En}}} 
& $\sigma^2$-fixed  & --    &  0.026  & 0.509  & 1.422 & 0.216 \\
& MCD  & 0.099 & 0.012  & 0.462  & 1.319 & 0.215  \\
  &  DE & 0.134 & 0.019  & \underline{0.366}  & 1.156 & \underline{0.460} \\
  & HTS  & 0.077 & 0.024  & 0.518  & 1.432 & 0.216 \\
  & HTS+MCD   & \underline{0.229} & \underline{0.006} & 0.502 & \underline{1.276} & 0.195\\
  & DUP & 0.082 & 0.024 & 0.516 & 1.418 & 0.216 \\
%   & DUP_{f_\mathrm{\{Q,MCD\}}}^{L_3} &  &  &  &  &   \\ 
  \midrule 
\multirow{6}{*}{\rotatebox{90}{\textsc{De-En}}} 
& $\sigma^2$-fixed  & --    & 0.025  & 0.403  & 1.398 & 0.573  \\
& MCD   & 0.044 & 0.030  & \underline{0.312}  & 1.343 & 0.568 \\
  &  DE & 0.056 & 0.030  & 0.321  & 1.374 & \underline{0.574} \\
  &  HTS  & 0.099 & 0.024  & 0.425  & 1.389 & 0.573 \\
  & HTS+MCD  & \underline{0.148} & \underline{0.014} & 0.463 & \underline{1.107} & 0.563 \\
  & DUP  & 0.100 & 0.023 & 0.432 & 1.382 & 0.573\\
%   & DUP_{f_\mathrm{\{Q,MCD\}}}^{L_3}  &  &  &  &  & \\ 
  \midrule
\multirow{6}{*}{\rotatebox{90}{\textsc{Ja-En}}}
& $\sigma^2$-fixed  & --    &  0.020  & 0.494  & 1.344 & 0.348 \\
& MCD  & 0.064 & 0.008  & 0.532  & \underline{1.280} & 0.349 \\
  &  DE & 0.079 & 0.012  & \underline{0.502}  & 1.305 & \underline{0.360} \\
  &  HTS  & 0.145 & 0.015  & 0.534  & 1.351 & 0.348 \\
  & HTS+MCD & \underline{0.215} & \underline{0.007} & 0.611 & 1.322 & 0.339 \\
  & DUP & 0.129 & 0.016 & 0.513 & 1.333 & 0.348 \\
%   & DUP_{f_\mathrm{\{Q,MCD\}}}^{L_3}  &  &  &  &  & \\ 
  \midrule
\multirow{6}{*}{\rotatebox{90}{\textsc{Km-En}}} 
& $\sigma^2$-fixed  & --    &  0.007  & 0.618  & 1.246 & 0.452 \\
& MCD  & 0.012 & 0.005  & 0.663  & 1.235 & 0.453 \\
  &  DE & 0.067 & \underline{0.003}  & \underline{0.631}  & \underline{1.226} & \underline{0.464} \\
  &  HTS  & \underline{0.147} & 0.004  & 0.661  & 1.255 & 0.452 \\
  & HTS+MCD  & 0.143 & 0.015 & 0.836 & 1.263 & 0.452  \\
  & DUP  & 0.144 & 0.004 & 0.638 & 1.239 & 0.452 \\
%   & DUP_{f_\mathrm{\{Q,MCD\}}}^{L_3}  &  &  &  &  & \\ 
  \midrule
\multirow{6}{*}{\rotatebox{90}{\textsc{Pl-En}}}
& $\sigma^2$-fixed  & --    &  0.027  & 0.586  & 1.518 & 0.264 \\
& MCD  & 0.063 & 0.029  & \underline{0.472}  & 1.450 & 0.265  \\
  &  DE & 0.029 & 0.029  & 0.500  & 1.490 & \underline{0.271} \\
  &  HTS  & 0.045 & 0.025  & 0.609  & 1.530 & 0.264 \\
  & HTS+MCD  & \underline{0.139} & \underline{0.008} & 0.502  & \underline{1.424} & 0.268 \\
  & DUP  & 0.048 & 0.025 & 0.604 & 1.519 & 0.264 \\
%   & DUP_{f_\mathrm{\{Q,MCD\}}}^{L_3}  &  &  &  &  & \\ 
  \midrule
\multirow{6}{*}{\rotatebox{90}{\textsc{Ps-En}}}
& $\sigma^2$-fixed  & --    &  0.005  & 0.735  & 1.319 & 0.325 \\
& MCD & 0.028 & 0.006  & 0.740  & \underline{1.291} & 0.327 \\
  &  DE & 0.040 & \underline{0.004}  & \underline{0.732}  & 1.295 & \underline{0.330} \\
  &  HTS  & 0.110 & 0.005  & 0.735  & 1.315 & 0.325 \\
  & HTS+MCD & \underline{0.111} & 0.013 & 0.849 & 1.315 & 0.297 \\
  & DUP  & 0.097 & 0.005 & \underline{0.732} & 1.317 & 0.325 \\
%   & DUP_{f_\mathrm{\{Q,MCD\}}}^{L_3}  &  &  &  &  & \\ 
  \midrule
\multirow{6}{*}{\rotatebox{90}{\textsc{Ru-En}}}
& $\sigma^2$-fixed  & --    & 0.031  & 0.454  & 1.574 & 0.288  \\
& MCD & 0.058 & 0.033  & \underline{0.373}  & 1.528 & 0.281  \\
  &  DE & 0.056 & 0.034  & 0.401  & 1.526 & \underline{0.300} \\
  &  HTS  & 0.087 & 0.030  & 0.464  & 1.575 & 0.288 \\
  & HTS+MCD  & \underline{0.161} & \underline{0.013} & 0.493 & \underline{1.520} & 0.209\\
  & DUP  & 0.073 & 0.029 & 0.467 & 1.570 & 0.288\\
%   & DUP_{f_\mathrm{\{Q,MCD\}}}^{L_3}  &  &  &  &  & \\ 
  \midrule
\multirow{6}{*}{\rotatebox{90}{\textsc{Ta-En}}}
& $\sigma^2$-fixed  & --    &  0.028  & 0.588  & 1.416 & 0.346 \\
& MCD  & 0.047 & 0.022  & \underline{0.567}  & 1.357 & 0.346 \\
  &  DE & 0.058 & 0.024  & 0.585  & 1.410 & \underline{0.357} \\
  &  HTS  & 0.081 & 0.023  & 0.577  & 1.468 & 0.346 \\
  & HTS+MCD   & \underline{0.250} & \underline{0.016} & 0.642 & \underline{1.300} & 0.284\\
  & DUP  & 0.079 & 0.025 & 0.602 & 1.420 & 0.346\\
%   & DUP_{f_\mathrm{\{Q,MCD\}}}^{L_3}  &  &  &  &  & \\ 
  \midrule
\multirow{6}{*}{\rotatebox{90}{\textsc{Zh-En}}}
& $\sigma^2$-fixed  & --    &  0.021  & 0.518  & 1.510 & 0.303 \\
& MCD  & 0.051 & 0.020  & \underline{0.447}  & 1.458 & 0.302 \\
  &  DE &  0.054 & 0.022  & 0.451  & 1.481 & \underline{0.310} \\
  &  HTS  & 0.082 & 0.020  & 0.533  & 1.508 & 0.303 \\
  & HTS+MCD  & \underline{0.186} & \underline{0.006} & 0.504 & \underline{1.377} & 0.278 \\
  & DUP  & 0.089 & 0.020 & 0.523 & 1.502 & 0.303 \\\midrule
  \multirow{6}{*}{\rotatebox{90}{\textsc{Xx-En}}} 
  & $\sigma^2$-fixed  & --    & 0.023  & 0.529  & 1.448 & 0.334 \\
& MCD    & 0.055 & 0.020  & \underline{0.477}  & 1.392 & 0.332\\
  &  DE  & 0.053 & 0.022  & 0.480  & 1.418 & \underline{0.342} \\
  & HTS  & 0.090 & 0.021  & 0.546  & 1.455 & 0.334 \\
  & HTS+MCD &	\underline{0.176} &	\underline{0.011} &	0.600 &	\underline{1.323} & 0.276 \\
  & DUP  & 0.089 &	0.021 &	0.542 &	1.442 &	0.334\\
%   & DUP_{f_\mathrm{\{Q,MCD\}}}^{L_3}  &  &  &  &  & \\ 
\bottomrule
\end{tabular}
}
\addtolength{\tabcolsep}{-0.5pt}
\caption{Results for segment-level DA prediction for Xx-En LPs. \underline{Underlined} numbers indicate the best result for each evaluation metric in each language pair.}
\label{tab:da_notqe_xxen}
\end{table}

% Results for non-english language pairs from WMT20 dataset are presented in Table \ref{tab:table_notqe_wmt20_noneng}
% \input{table_notqe_wmt20_xxxx}

\section{MQM experiments}
\label{sec:app_mqm_experiments}
We provide extended results for each language pair in the MQM 2021 test set in Table~\ref{tab:mqm_ft}.

\begin{table}[t]
\small
\centering
\addtolength{\tabcolsep}{-0.5pt}
\resizebox{7.7cm}{!}{
\begin{tabular}{clccccc}
\toprule
&  & UPS $\uparrow$   & ECE $\downarrow$ & Sha. $\downarrow$ & NLL $\downarrow$ & PPS $\uparrow$ \\  
\midrule
\multirow{7}{*}{\rotatebox{90}{\textsc{En-De}}} 
& $\sigma^2$-fixed & — & 0.053  & 0.228  & 2.543 & 0.342 \\ 
& MCD & 0.132 & \underline{0.026} & 0.228 & 1.984 & 0.391 \\
& DE & 0.075 & 0.057 & \underline{0.155} & 2.911 & \underline{0.422} \\
  & HTS  & 0.236  & 0.029  & 0.192  & 2.274  & 0.370  \\
& HTS+MCD  & 0.232  & \underline{0.025}  & 0.280  & \underline{1.841}  & 0.365  \\
& KL  & \underline{0.251}  & 0.052  & \underline{0.168}  & 2.641  & 0.391  \\
& DUP  & 0.186  & 0.051  & 0.273  & 2.215  & 0.342  \\
  %   HTS + DEUP + Features (Trained on WMT20*0.7)
  %& DUP_{f_\mathrm{\{Q,MCD\}}}^{L_3} &   \\ 
  \midrule 
\multirow{7}{*}{\rotatebox{90}{\textsc{Zh-En}}}  
& $\sigma^2$-fixed & — & 0.058  & 0.516  & 1.611 & 0.439 \\
  & MCD & 0.253 & \underline{0.009} & 0.500 & \underline{1.219} & 0.590 \\
& DE & 0.157 & 0.045 & \underline{0.420} & 1.381 & \underline{0.601} \\
  & HTS  & 0.365  & 0.036  & 0.514  & 1.338  & 0.564  \\
& HTS+MCD  & 0.380  & 0.035  & 0.566  & 1.248  & 0.570  \\
& KL  & 0.348  & 0.037  & 0.515  & 1.353  & 0.547  \\
& DUP  & \underline{0.386}  & 0.039  & 0.949  & 1.419  & 0.439  \\%   HTS + DEUP + Features (Trained on WMT20*0.7)
  %& DUP_{f_\mathrm{\{Q,MCD\}}}^{L_3} &  &  &  &  &   \\ 
  \midrule 
\multirow{7}{*}{\rotatebox{90}{\textsc{En-Ru}}}  
& $\sigma^2$-fixed & — & 0.053  & 0.338  & 2.217 & 0.340 \\
& MCD & 0.136 & 0.039 & 0.243 & 1.944 & 0.376 \\
& DE & 0.144 & 0.052 & \underline{0.100} & 3.803 & \underline{0.392} \\
 & HTS  & \underline{0.308}  & 0.059  & \underline{0.104}  & 3.317  & 0.377  \\
& HTS+MCD  & 0.304  & 0.049  & 0.285  & 1.822  & 0.375  \\
& KL  & 0.279  & 0.050  & 0.095  & 3.976  & 0.372  \\
& DUP  & 0.260  & \underline{0.029}  & 0.608  & \underline{1.783}  & 0.340  \\
%   HTS + DEUP + Features (Trained on WMT20*0.7)
  %& DUP_{f_\mathrm{\{Q,MCD\}}}^{L_3} &  &  &  &  &  \\ 
  \midrule
\multirow{7}{*}{\rotatebox{90}{\textsc{Avg}}}  
& $\sigma^2$-fixed & — & 0.055  & 0.371  & 2.090 & 0.377 \\
& MCD & 0.179 & \underline{0.024} & 0.334 & 1.686 & 0.460 \\
& DE & 0.128 & 0.051 & \underline{0.236} & 2.631 & \underline{0.479} \\
& HTS  & 0.307  & 0.041  & 0.284  & 2.264  & 0.445  \\
& HTS+MCD  & \underline{0.311}  & 0.037  & 0.388  & \underline{1.614}  & 0.445  \\
& KL  & 0.296  & 0.046  & 0.273  & 2.595  & 0.443  \\
& DUP  & 0.285  & 0.039  & 0.634  & 1.778  & 0.377  \\ %   HTS + DEUP + Features (Trained on WMT20*0.7)
  %& DUP_{f_\mathrm{\{Q,MCD\}}}^{L_3} &  &  &  &  &  \\
\bottomrule
\end{tabular}
}
\addtolength{\tabcolsep}{-0.5pt}
\caption{Results for segment-level MQM predictions with fine-tuning on MQM 2020 data. \underline{Underlined} numbers indicate the best result for each evaluation metric in each language pair.}
\label{tab:mqm_ft}
\end{table}

We also present results without fine-tuning on the MQM data in Table~\ref{tab:mqm}, to facilitate comparisons. For these experiments we use the models trained on the WMT DA data (performance for these models is also reported in Tables~\ref{tab:da_notqe_enxx} and \ref{tab:da_notqe_xxen}). We can see that without further finetuning on MQM scores all models with the exception of the ones based on variance (\textbf{MCD} and \textbf{DE}) have a significant drop in performance. %Thus approaches that require supervision in order to learn uncertainty related parameters are more sensitive to changes of evaluation schemas 

\begin{table}[ht!]
\small
\centering
\addtolength{\tabcolsep}{-0.5pt}
\resizebox{7.7cm}{!}{
\begin{tabular}{clccccc}
\toprule
&  & UPS $\uparrow$  & ECE $\downarrow$ & Sha. $\downarrow$ & NLL $\downarrow$  & PPS $\uparrow$ \\  
\midrule
\multirow{5}{*}{\rotatebox{90}{\textsc{En-De}}} 
% & Baseline & 0.302 & - & 1.602 & 0.051 & 2.625 \\
& MCD  & \underline{0.134} & 0.069 & 1.019  & 0.577 & 0.295\\
  &  DE  & 0.104 & \underline{0.021} & 1.03 & 0.644 & \underline{0.332} \\
  & HTS & 0.094 & \underline{0.039} & 0.274  & 2.567 & 0.326  \\
  & HTS + MCD & 0.126 & \underline{0.021} & 0.356 & 1.502 & 0.291 \\
  & DUP & 0.038& 0.054 & \underline{0.241}  & 2.248 & 0.302 \\ 
%   & DUP_{f_\mathrm{\{Q,MCD\}}}^{L_3} & 0.302 & 0.018 & 2.241 & 0.054 & 0.243   \\ 
  \midrule %CZ to add
  \multirow{5}{*}{\rotatebox{90}{\textsc{Zh-En}}}  
%   & Baseline & 0.434 & - & 1.820 & 0.045 & 4.667  \\
  & MCD   & 0.115 & 0.081 & 1.321  & 0.956 & 0.441\\
  &  DE   & 0.14  & 0.025 & 1.143 & 0.911 & \underline{0.457}\\
  & HTS   & 0.082 & \underline{0.013} & 0.595 & 1.615 & 0.436\\
  & HTS + MCD  & -0.006  & \underline{0.013} & 0.637 & 1.42 & 0.433\\
  & DUP  & \underline{0.17}  & 0.05 & \underline{0.469} & 1.814 & 0.434\\ %   HTS + DEUP + Features (Trained on WMT20*0.7)
%   & DUP_{f_\mathrm{\{Q,MCD\}}}^{L_3} & 0.434 & 0.035 & 1.812 & 0.051 & 0.486  \\ 
  \midrule %CZ to add
\multirow{5}{*}{\rotatebox{90}{\textsc{En-Ru}}}  
% & Baseline  & 0.290 & - & 1.995 & 0.028 & 7.089 \\
& MCD    & 0.14  & 0.069 & 1.242 & 0.563 & 0.306\\
  &  DE   & 0.117  & 0.078 & 1.332 & 0.684 & 0.318\\
  & HTS  & 0.134  & 2.035 & \underline{0.306}  & \underline{0.021} & \underline{0.337}\\
  & HTS + MCD  & -0.042  & \underline{0.016} & 0.459 & 1.492 & 0.333\\
  & DUP  & \underline{0.139}  & 0.045 & 0.35 & 2.238 & 0.290\\ %   HTS + DEUP + Features (Trained on WMT20*0.7)
%   & DUP_{f_\mathrm{\{Q,MCD\}}}^{L_3} & 0.290 & 0.120 & 2.249 & 0.045 & 0.352  \\ 
  \midrule
\multirow{5}{*}{\rotatebox{90}{\textsc{AVG}}}  
% & Baseline & & & & & \\
& MCD   & 0.356 & \underline{0.129} & 0.722 & 0.074 & 1.215 \\
  &  DE  & \underline{0.377} & 0.123 & 0.763 & 0.042 & 1.179 \\
  & HTS & 0.289 & 0.079 & \underline{0.012} & 1.34 & 0.341  \\
  & HTS + MCD & 0.286 & -0.017 & 1.076 & \underline{0.011} & 0.41 \\ 
  & DUP & 0.272 & 0.115 & 1.489 & 0.035 & \underline{0.306} \\ 
%   & DUP_{f_\mathrm{\{Q,MCD\}}}^{L_3} &  &  &  &  &  \\
\bottomrule
\end{tabular}
}
\addtolength{\tabcolsep}{-0.5pt}
\caption{Results for segment-level MQM prediction without fine-tuning. \underline{Underlined} numbers indicate the best result for each evaluation metric in each language pair.}
% \andre{Excluding En-Ru, these results show that (1) HTS and DEUP are better than the estimates based on model variance (MCD and DE) in UPS, ECE, Sharpness, and worse in NLL. The En-Ru UPS is super-weird. Do we have an intuition about why there  are so many negatives? are those annotations reliable? we should consider excluding that LP if we don't trust the annotations. It looks like the problem still exists in the fine-tuned model (Tab 4). Also, are we able to add KL to this table as we have in Tab 4? Or it's only possible to consider KL in the fine-tuned model setting?}
\label{tab:mqm}
\end{table}

% For calibration related indicators higher scores indicate lower performances. For correlation related indicators, higher scores indicate better performance. For each language pair the best values of each indicator are bold.

\section{Model implementation and parameters}
\label{sec:model_params}

\begin{table*}[ht!]
\small
\centering
%\resizebox{\columnwidth}{!}{
\begin{tabular}{lccc}
\toprule 
\textbf{Hyperparameter}  & \textbf{MCD/DE/Vanilla COMET}  & \textbf{HTS/KL}  & \textbf{DUP}  \tabularnewline
\midrule 
Encoder Model  & XLM-R (large)  & XLM-R (large) & XLM-R (large) \tabularnewline
Optimizer  & Adam  & Adam & Adam \tabularnewline
No. frozen epochs  & 0.3  & 0.3 & 0.3 \tabularnewline
Learning rate  & 3e-05  & 3e-05 & 3e-05 \tabularnewline
Encoder Learning Rate  & 1e-05  & 1e-05 & 1e-05 \tabularnewline  %check 
Layerwise Decay  & 0.95  & 0.95 & 0.95 \tabularnewline
Batch size  & 4  & 4 & 4 \tabularnewline %check 
Loss function  & Mean squared error  & $\mathcal{L}_{\mathrm{HTS}}$ / $\mathcal{L}_{\mathrm{KL}}$ &  $\mathcal{L}^\mathrm{E}_{\mathrm{HTS}}$ [$\mathcal{L}^\mathrm{E}_{\mathrm{ABS}}$ / $\mathcal{L}^\mathrm{E}_{\mathrm{SQ}}$] \tabularnewline
Dropout  & 0.15  & 0.15 & 0.15 \tabularnewline
Hidden sizes  & [3072, 1024]  & [3072, 1024] & [3072, 1024] \tabularnewline
Encoder Embedding layer & Frozen & Frozen & Frozen \tabularnewline
Bottleneck layer size & - & - & 256 \tabularnewline
FP precision  & 32  & 32 & 32 \tabularnewline
No. Epochs (training) & 2 & 2 & 2 \\
No. Epochs (fine-tuning) & 1 & 1 & 1 \\
\bottomrule
\end{tabular}
\caption{Hyperparameters used to train uncertainty prediction methods.} 
\label{tab:hp}
%}
\end{table*}

Table \ref{tab:hp} shows the hyperparameters used to train the following uncertainty prediction models: \textbf{MCD}, \textbf{DE}, \textbf{HTS}, \textbf{KL} and \textbf{DUP}. For deep ensembles we trained 4 models with different seeds and as a fifth model we used the \textit{wmt-comet-da} available at \url{https://github.com/Unbabel/COMET} (in the table we refer to it as \textbf{Vanilla \textsc{Comet}}).

% Each of these models has 583M parameters and were trained on a single Nvidia Quadro RTX 8000 GPU\footnote{\url{https://www.nvidia.com/en-us/design-visualization/quadro/rtx-8000/}} for $\approx34$ and $\approx3.5$ hours for the DA models and HTER models, respectively. 
% Regarding the validation performance recorded during training, the DA models achieve a PPS of $0.612\pm0.002$, while the HTER models achieve a PPS of $0.663\pm0.012$.

\section{Performance indicators}
\label{sec:app_metrics}

We briefly describe below each of the metrics reported for the experiments of this paper, provide the formulas for each one and the motivation for using them. 
For all described metrics we assume access to a test set $\mathcal{D} = \{\langle s_j, t_j, \mathcal{R}_j, q_j^* \rangle\}_{j=1}^{|\mathcal{D}|}$, consisting of samples paired with their ground truth quality scores.

\paragraph{Calibration Error} 
To estimate how well-calibrated the methods are we compute expected calibration error (ECE; \citealt{naeini2015obtaining, kuleshov2018accurate}), which is defined as:
\begin{align}
\label{eq:ece}
    \mathrm{ECE} &= \frac{1}{M}\sum_{b=1}^{M} |\mathrm{acc}(\gamma_{b}) - \gamma_{b}|,
\end{align}
where each $b$ is a bin representing a confidence level $\gamma_b$, and $\mathrm{acc}(\gamma_{b})$ is the fraction of times the ground truth $q^*$ falls inside the confidence interval $I(\gamma_b)$:
\begin{equation}
\label{eq:ece_acc}
    \mathrm{acc}(\gamma_{b}) = \frac{1}{|\mathcal{D}|} \sum_{\langle s, t, \mathcal{R}, q^* \rangle \in \mathcal{D}} \mathds{1}(q^* \in I(\gamma_b)).
\end{equation}
We use this metric with $M=100$, similarly to previous works.

\paragraph{Negative log-likelihood} 

The negative log-likelihood (NLL) captures both accuracy- and uncertainty-related performance, since it essentially considers the log-likelihood of the true quality score $q^*$ based on the distribution estimated by the predicted variance (uncertainty). Thus it penalizes predictions that are accurate but have too high uncertainty (since they will become flat distributions with low probability everywhere), and even more severely incorrect predictions with high confidence, but is more lenient with predictions that are inaccurate but have high uncertainty.

\begin{equation}
    \mathrm{NLL} = -\frac{1}{|\mathcal{D}|}\sum_{\langle s, t, \mathcal{R}, q^* \rangle \in \mathcal{D}} \log \hat{p}(q^* \mid \langle s, t, \mathcal{R} \rangle).
\end{equation}

Note that it is possible to calculate the optimal fixed variance that minimizes NLL by: 
\begin{equation}
    \label{eq:nll_opt}
    \sigma_{\mathrm{fixed}}^2 = \frac{1}{|\mathcal{D}|} \sum_{j = 1}^{|\mathcal{D}|} (q_j^* - \hat{\mu}_j)^2.
\end{equation}

\paragraph{Sharpness} 
% The metrics above do not sufficiently account for how ``tight'' the uncertainty interval is around the predicted value, and thus might generally favour predictors that produce wide and uninformative confidence intervals. 
% To guarantee useful uncertainty estimation, confidence intervals should not only be calibrated, but also sharp.
To ensure informative uncertainty estimation, confidence intervals should not only be calibrated, but also sharp.
We measure sharpness using the predicted variance $\hat{\sigma}^2$, as defined in \citet{kuleshov2018accurate}:

\begin{equation}
    \mathrm{sha}(\hat{p}_Q) = \frac{1}{|\mathcal{D}|} \sum_{\langle s, t, \mathcal{R} \rangle \in \mathcal{D}}  \hat{\sigma}^2.
\end{equation}

\paragraph{Pearson correlations} 
\label{sec:pearson_correlations}

The \textbf{predictive Pearson score} (PPS), evaluates the predictive accuracy of the system -- it is the Pearson correlation $r(q^*, \hat{q})$ between the ground truth quality scores $q^*$ and the system predictions $\hat{q}$ in the dataset $\mathcal{D}$. 
The \textbf{uncertainty Pearson score} (UPS) $r(|q^*-\hat{q}|, \hat{\sigma})$, measures the alignment between the prediction errors $|q^* - \hat{q}|$ and the uncertainty estimates $\hat{\sigma}$.

\section{Uncertainty on OOD examples}
\label{sec:app_ood_ext}

\begin{figure*}[hbt!]
\centering
\begin{subfigure}{0.45\textwidth}
\includegraphics[width=\textwidth]{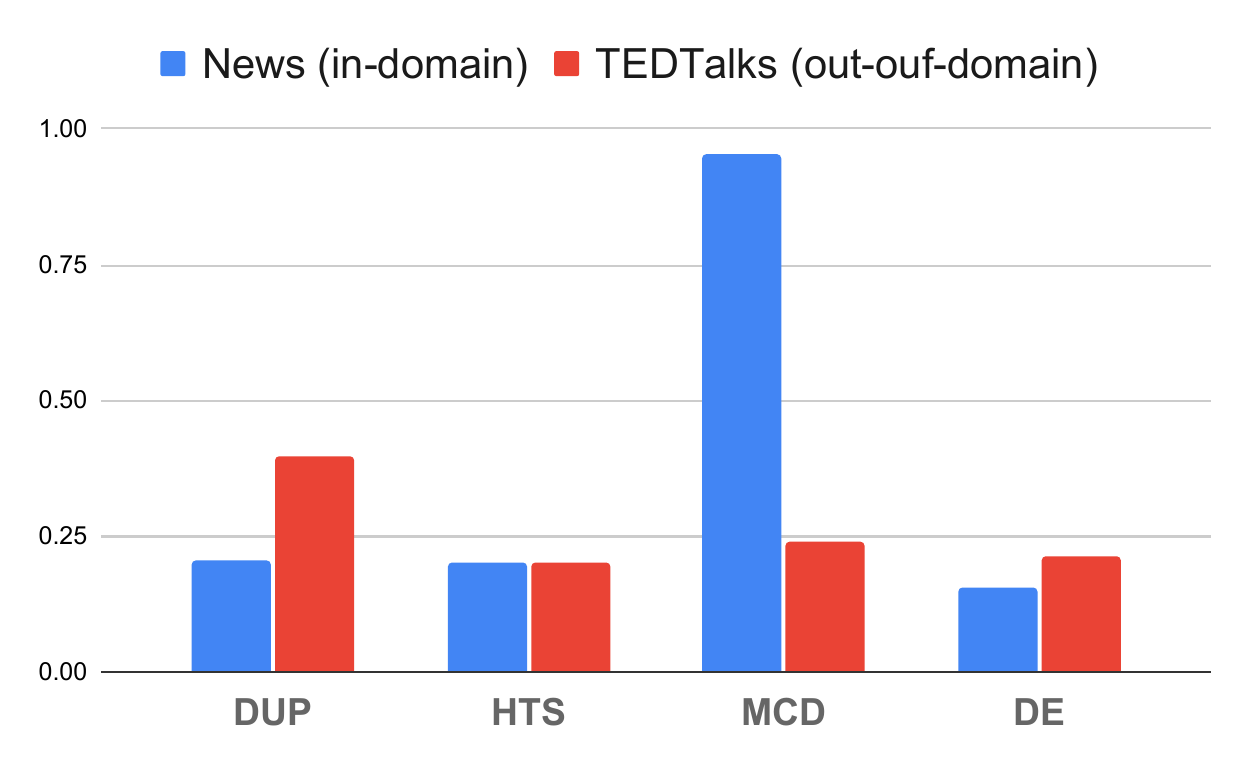}
\subcaption{\textsc{En-De}}
\end{subfigure}
\begin{subfigure}{0.45\textwidth}
\includegraphics[width=\textwidth]{images/enru-sharpness.pdf}
\subcaption{\textsc{En-Ru}}
\end{subfigure}
\begin{subfigure}{0.45\textwidth}
\includegraphics[width=\textwidth]{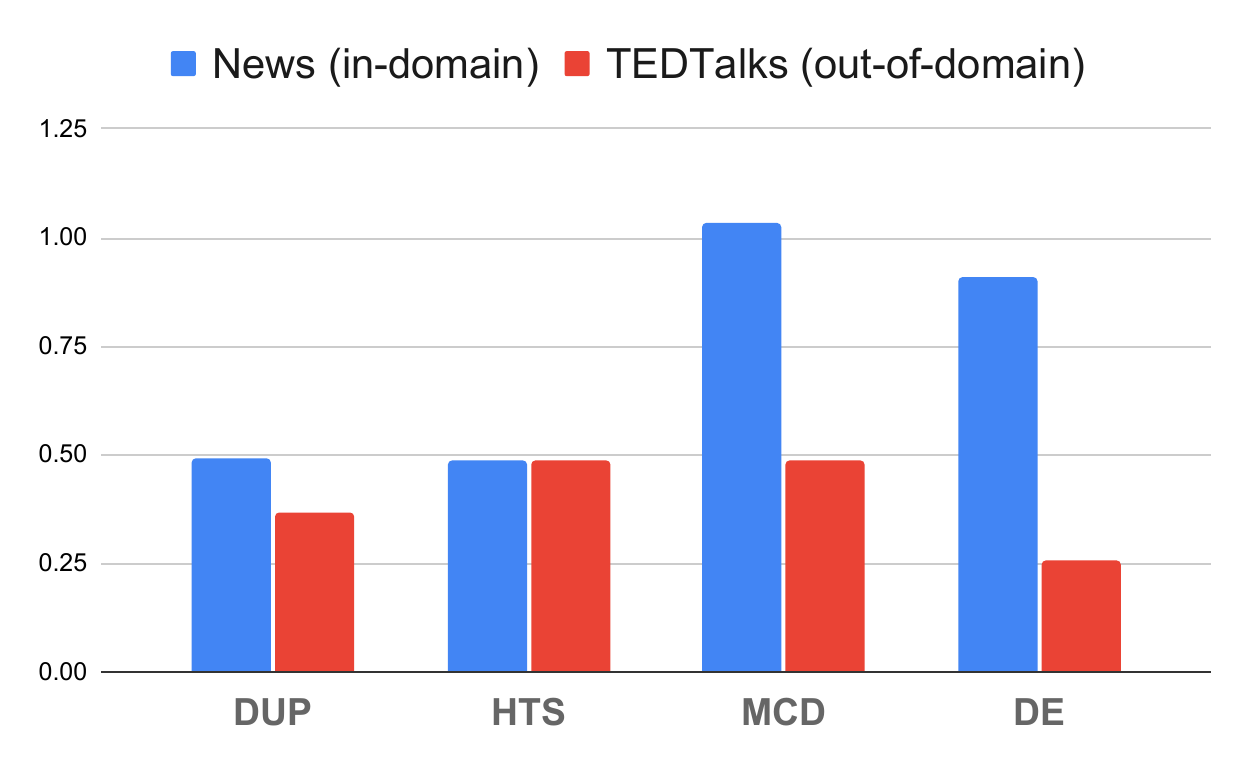}
\subcaption{\textsc{Zh-En}}
\end{subfigure}
\caption{Sharpness for in-domain \color{blue}(blue) \color{black} News WMT21 MQM data and out-of-domain \color{red}(red) \color{black} TEDTalks WMT21 MQM data. We show changes in sharpness values on each language pair separately, for the \textbf{DUP}, \textbf{HTS}, \textbf{MCD} and \textbf{DE} models finetuned on News WMT20 MQM data.  }
\label{fig:sharpness_ext}
\end{figure*}

We provide the comparison of the sharpness value, representing the quantified uncertainty for in-domain (ID) data (WMT21 news data with MQM annotations) and out-of-domain (OOD) data (WMT21 TEDTalks data with MQM annotations) in Figure \ref{fig:sharpness_ext}. Sharpness as explained in App. \ref{sec:app_metrics}, is an indicator of the overall estimated confidence of a model over a given dataset. Thus we want to examine whether the estimated confidence intervals for the OOD data are representative of the expected increase in epistemic uncertainty. 

Looking at the sharpness variation per language pair, we can see that for \textsc{En-De} and \textsc{En-Ru}, where the aleatoric uncertainty is relatively low as indicated by the low HTS values, the sharpness increases significantly for the DUP model. This behaviour however does not hold for cases where aleatoric uncertainty is higher (\textsc{Zh-En}). We speculate that this could be attributed to the fact that DUP is trained to capture total uncertainty, instead of only epistemic, and thus it is sensitive to increased noise in the data. Further experiments would be needed to verify this hypothesis.

Across language pairs, the values for HTS remain the same for ID and OOD, while for MCD we have the opposite effect than what was expected: sharpness drops significantly for OOD data in all language pairs. This further supports our claim that uncertainty predictors relying on model variance are not optimal to represent epistemic uncertainty. 

\edit{
For completeness we also provide the results for the rest of performance indicators on the TedTalk dats in Table \ref{tab:ted_full}. Note that for the OOD experiments we sampled half the dataset for testing and reserved the rest for calibration (resulting in approx. 4K segments per language pair for each split).}

\begin{table}[t]
\small
\centering
\addtolength{\tabcolsep}{-0.5pt}
\resizebox{7.7cm}{!}{
\begin{tabular}{clccccc}
\toprule
&  & UPS $\uparrow$   & ECE $\downarrow$ & Sha. $\downarrow$ & NLL $\downarrow$ & PPS $\uparrow$ \\  
\midrule
\multirow{6}{*}{\rotatebox{90}{\textsc{En-De}}}  
& $\sigma^2$-fixed  & --    & 0.072 & \underline{0.146} & 2.957 & 0.526 \\
& MCD     & 0.178 & \underline{0.052} & 0.147 & 2.500 & 0.540 \\
& DE      & 0.371 & 0.062 & 0.314 & 1.977 & \underline{0.571} \\
& HTS     & 0.290 & 0.070 & 0.251 & 2.239 & 0.425 \\
& HTS+MCD & \underline{0.401} & 0.073 & 0.227 & \underline{1.756} & 0.545 \\
& DE      & 0.346 & 0.058 & 0.346 & 2.219 & 0.526\\\midrule
\multirow{6}{*}{\rotatebox{90}{\textsc{En-Ru}}}  
& $\sigma^2$-fixed  & --    & 0.057 & 0.229 & 2.095 & 0.436 \\
& MCD     & 0.086 & 0.065 & 0.238 & 1.846 & 0.425 \\
& DE      & 0.271 & 0.057 & 0.346 & 1.679 & \underline{0.441} \\
& HTS     & 0.267 & 0.084 & \underline{0.151} & 2.506 & 0.372 \\
& HTS+MCD & \underline{0.293} & 0.068 & 0.402 & \underline{1.473} & 0.387 \\
& DUP     & 0.282 & \underline{0.047} & 0.300 & 1.781 & 0.436 \\\midrule
\multirow{6}{*}{\rotatebox{90}{\textsc{Zh-En}}}  
& $\sigma^2$-fixed  & --    & 0.033 & 0.397 & 2.203 & 0.434 \\
& MCD     & 0.063 & \underline{0.023} & 0.283 & 2.348 & 0.447 \\
& DE      & 0.23  & 0.036 & 0.586 & 1.865 & 0.456 \\
& HTS     & \underline{0.378} & 0.067 & \underline{0.135} & 2.685 & \underline{0.544} \\
& HTS+MCD & 0.288 & 0.073 & 0.223 & 2.276 & 0.425 \\
& DUP     & 0.271 & 0.030 & 0.825 & \underline{1.718} & 0.434\\ \bottomrule
\end{tabular}
}
\addtolength{\tabcolsep}{-0.5pt}
\caption{Results for segment-level MQM predictions on TEDTalk data. \underline{Underlined} numbers indicate the best result for each evaluation metric in each language pair.}
\label{tab:ted_full}
\end{table}

% \begin{figure*}\centering
% \subfloat[\textsc{En-De}]{\label{a}\includegraphics[width=.45\textwidth]{images/sharp_ende.pdf}}\hfill
% \subfloat[\textsc{En-Ru}]{\label{b}\includegraphics[width=.45\textwidth]{images/sharp_enru.pdf}}\par 
% \subfloat[\textsc{Zh-En}]{\label{c}\includegraphics[width=.45\textwidth]{images/sharp_zhen.pdf}}
% \caption{Sharpness}
% \label{fig}
% \end{figure*}

\section{Ablation tests for DUP}
\label{sec:ablation}

We present different ablation tests on the DUP architecture to compare the impact of different modelling choices on the training of the model. Our ablation tests are focusing on the second step model, $\mathcal{M}_\mathrm{E}$, since it is the one that accounts for the uncertainty predictions.

\subsection{Comparison of loss functions}
We explore three different loss functions for the $\mathcal{M}_\mathrm{E}$ model of \textbf{DUP}, described in Eqs. \ref{eq:loss1}--\ref{eq:loss31}.

\begin{align}
   \mathcal{L}^\mathrm{E}_{\mathrm{ABS}}(\hat{\epsilon}; \epsilon^*) &= (\epsilon^* - \hat{\epsilon})^2\label{eq:loss1}\\
   \mathcal{L}^\mathrm{E}_{\mathrm{SQ}}(\hat{\epsilon}; \epsilon^*) &= ((\epsilon^*)^2 - \hat{\epsilon}^2)^2 \label{eq:loss2}\\
   \mathcal{L}^\mathrm{E}_{\mathrm{HTS}}(\hat{\epsilon}; \epsilon^*) &= \frac{(\epsilon^*)^2}{2 \hat{\epsilon}^2} + \frac{1}{2}\log(\hat{\epsilon})^2.
   \label{eq:loss31}
\end{align} 

Losses $\mathcal{L}^\mathrm{E}_\mathrm{ABS}$ and $\mathcal{L}^\mathrm{E}_\mathrm{SQ}$ are variations of the mean squared error loss, using as argument either the absolute error $\hat{\epsilon}$ or the squared error $\hat{\epsilon}^2$.

We compare the performance of DUP models trained using the different losses on the segment-level DA data. According to the results in Table~\ref{tab:da_losses}, all three losses perform similarly, with a slight advantage to $\mathcal{L}_\mathrm{HTS}^\mathrm{E}$. 
This motivated our choice to run the experiments discussed in the main paper using this loss as a representative of \textbf{DUP}. %\andre{we're still missing the second loss, will we have it on time?}
%\andre{in case we have results for each LP in the appendix, mention it here}

\begin{table}[ht!]
\small
\centering
\addtolength{\tabcolsep}{-0.5pt}
\resizebox{7.7cm}{!}{
\begin{tabular}{clccccc}
\toprule
 &  & UPS $\uparrow$  & ECE $\downarrow$ & Sha. $\downarrow$ & NLL $\downarrow$ & PPS $\uparrow$\\  
\midrule
\multirow{3}{*}{\rotatebox{90}{\textsc{En-Xx}}} 
  & DUP ${\mathcal{L}^\mathrm{E}_{\mathrm{ABS}}}$  & 0.134  & 0.013 & 0.295 & 1.019& 0.633 \\ [3pt]
  & DUP ${\mathcal{L}^\mathrm{E}_{\mathrm{SQ}}}$  &	0.140 &	0.012 &	0.315 &	1.022 & 0.633\\ [3pt] 
  & DUP ${\mathcal{L}^\mathrm{E}_{\mathrm{HTS}}}$  & 0.146  & 0.014 & 0.293 & 1.021 &  0.633 \\ 
  \midrule
\multirow{3}{*}{\rotatebox{90}{\textsc{Xx-En}}}  
  & DUP ${\mathcal{L}^\mathrm{E}_{\mathrm{ABS}}}$   & 0.081 & 0.017 & 0.527 & 1.471 & 0.287 \\[3pt]
  & DUP ${\mathcal{L}^\mathrm{E}_{\mathrm{SQ}}}$  & 0.084  &	0.017 &	0.534 &	1.470 & 0.287\\ [3pt]
  & DUP ${\mathcal{L}^\mathrm{E}_{\mathrm{HTS}}}$  &	0.086  &	0.017 &	0.524 &	1.473 & 0.287\\ \midrule
\multirow{3}{*}{\rotatebox{90}{\textsc{Avg}}} 
  & DUP ${\mathcal{L}^\mathrm{E}_{\mathrm{ABS}}}$   & 0.104 & 1.265 & 0.015 & 0.414 & 0.446\\ [3pt]
  & DUP ${\mathcal{L}^\mathrm{E}_{\mathrm{SQ}}}$ &	0.108 &	1.262 &	0.014 &	0.427\\ [3pt]
  & DUP ${\mathcal{L}^\mathrm{E}_{\mathrm{HTS}}}$  &	0.112 &	1.266 &	0.015 &	0.411 & 0.446\\ 
\bottomrule
\end{tabular}
}
\addtolength{\tabcolsep}{-0.5pt}
\caption{Comparison of different losses for the DUP method in segment-level DA prediction.}
\label{tab:da_losses}
\end{table}

% For calibration related indicators higher scores indicate lower performances. For correlation related indicators, higher scores indicate better performance. For each language pair the best values of each indicator are bold.

\subsection{Comparison of parameter sharing settings}

\begin{figure*}[ht!]
    \centering
    \includegraphics[width=\textwidth]{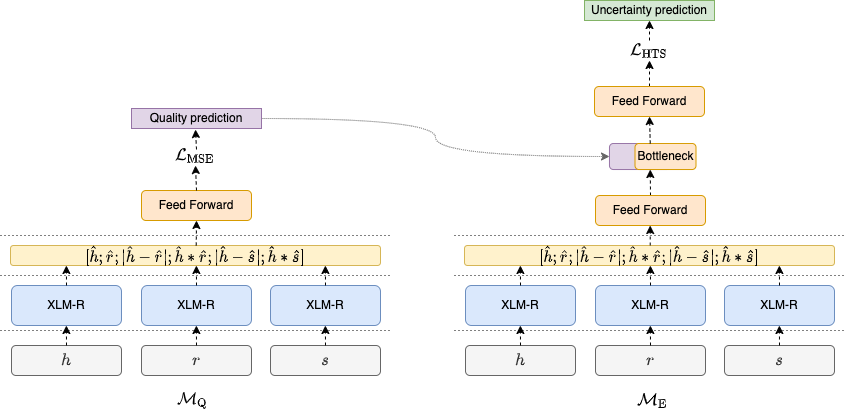}
    \caption{Architecture and dependencies of DUP $\mathcal{M}_\mathrm{Q}$ and $\mathcal{M}_\mathrm{E}$ models}
    \label{fig:arch}
\end{figure*}

For this paper the models used for $\mathcal{M}_\mathrm{Q}$ and $\mathcal{M}_\mathrm{E}$ use very similar architectures, except for the bottleneck layer, as depicted in Figure \ref{fig:arch}. We thus compare the impact of three different settings:
\begin{enumerate}
    \item \textbf{NS}: Not sharing any parameters and training $\mathcal{M}_\mathrm{E}$ from scratch.
    \item \textbf{S}: Sharing all (common) parameters between $\mathcal{M}_\mathrm{Q}$ and $\mathcal{M}_\mathrm{E}$; then keep fine-tuning $\mathcal{M}_\mathrm{E}$ on the new uncertainty (error) prediction task.
    \item \textbf{SF}: Sharing all (common) parameters between $\mathcal{M}_\mathrm{Q}$ and $\mathcal{M}_\mathrm{E}$ and freeze the XLM-R encoder weights and embeddings; then keep fine-tuning the rest of the  $\mathcal{M}_\mathrm{E}$ parameters on the new uncertainty (error) prediction task.
\end{enumerate}

\begin{table}[ht!]
\small
\centering
\addtolength{\tabcolsep}{-0.5pt}
\resizebox{7.7cm}{!}{
\begin{tabular}{clccccc}
\toprule
 & & UPS $\uparrow$ & ECE $\downarrow$ & Sha. $\downarrow$ & NLL $\downarrow$ & PPS $\uparrow$ \\  
\midrule
\multirow{3}{*}{\rotatebox{90}{\textsc{En-Xx}}} 
  & NS & 0.272 & 0.008  & 0.344  & 0.919 & 0.566\\ 
  & S & \underline{0.285} &	\underline{0.007} &	\underline{0.320} &	\underline{0.910} &	0.566\\ 
  & SF & 0.276 & 0.008 & 0.341 & 0.917 & 0.566  \\ 
  \midrule
\multirow{3}{*}{\rotatebox{90}{\textsc{Xx-En}}}  
  & NS &  0.090 & 0.021  & 0.546  & 1.455 & 0.334\\
  & S & 0.089 &	0.021 &	\underline{0.542} &	\underline{1.442} &	0.334\\ 
  & SF & \underline{0.093} & \underline{0.020} & 0.550 & 1.462 & 0.334 \\ \midrule
\multirow{3}{*}{\rotatebox{90}{\textsc{Avg}}} 
  & NS & 0.177 & 0.015  & 0.450  & 1.201 & 0.444\\
  & S & \underline{0.182} &	\underline{0.014} & \underline{0.437} &	\underline{1.190} &	0.444\\
  & SF & 0.180 & 0.015 & 0.451 & 1.204 & 0.444 \\
\bottomrule
\end{tabular}
}
\addtolength{\tabcolsep}{-0.5pt}
\caption{Comparison of different parameter share configurations for DUP.}
\label{tab:da_param}
\end{table}

The results are presented in Table \ref{tab:da_param}. We see that sharing parameters (\textbf{S}, \textbf{SF} settings) consistently results in a small boost for all uncertainty indicators. Since we do not see a significant further improvement by keeping the encoder frozen (\textbf{SF}), we perform the rest of the experiments presented in this work by simply sharing the parameters between $\mathcal{M}_\mathrm{Q}$ and $\mathcal{M}_\mathrm{E}$ (\textbf{S} setting).

\section{Results on other metrics}
\label{app:sec:other_metrics}

In this section we present results on the WMT 20 DA dataset using trainable metrics that differ to the \textsc{Comet} architecture, as an additional comparison. We select \textsc{Bleurt} and UniTE for this comparison. \textsc{Bleurt} \cite{sellam-etal-2020-learning}, is a multilingual metric with high performance, which unlike COMET jointly encodes only the translation and reference inputs in order to predict the quality score of a segment. UniTE is a newly proposed architecture \cite{wan2022unite} which is taking into account three different input combinations with the translation segment, namely reference-only, source-only and source-reference-combined. Note that we do not optimise the hyper-parameters of these metrics since we are only interested in comparing the overall behaviour. Hence, improved results could be expected upon optimisation.

We present results on \textsc{Bleurt} in Tables \ref{tab:da_bleurt_all}, \ref{tab:da_bleurt_enxx} and \ref{tab:da_bleurt_xxen}. We used a \textsc{Bleurt} implementation with RemBERT encoder \cite{chung2021rethinking}, trained on the same DA setup described in \S\ref{sec:experiments} of the main paper. We notice that we observe similar behaviour of the proposed uncertainty predictors to the one identified for COMET, with the exception of the heteroscedastic predictors (\textbf{HTS}, \textbf{HTS+MCD}). It seems that without access to the source segments it is harder for the heteroscedastic approach to learn to predict meaningful variance intervals. In other words, it seems to be harder for HTS approaches to identify noisy inputs relying only on the reference segments. This finding highlights the importance of including source segments towards identification of noisy inputs and prediction of segment level quality with higher confidence. In comparison, we can see that the \textbf{DUP} approach significantly improves the uncertainty correlation (UPS).

For UniTE, we show results in Tables \ref{tab:da_unite_avg}, \ref{tab:da_unite_enxx} and \ref{tab:da_unite_xxen}. We implemented UniTE with an InfoXLM encoder \cite{chi2021infoxlm}, trained on the same DA setup described in \S\ref{sec:experiments} of the main paper. We noticed that on average the correlations achieved for the UPS performance indicator are lower to the ones obtained with \textsc{Comet} and \textsc{Bleurt}, especially for \textbf{MCD}. However, they follow similar pattern to the one identified in the main paper: we obtain significantly better correlations both for the HTS and the DUP predictors.

\begin{table}[ht]
\small
\centering
\addtolength{\tabcolsep}{-0.5pt}
\resizebox{7.7cm}{!}{
\begin{tabular}{clccccc}
\toprule
&  & UPS $\uparrow$ & ECE $\downarrow$ & Sha. $\downarrow$   & NLL $\downarrow$ & PPS $\uparrow$ \\  
\midrule
\multirow{6}{*}{\rotatebox{90}{\textsc{En-Xx}}}  
  & $\sigma^2$-fixed & -- &  0.107  & 4.698  & 1.984 & 0.317 \\
  & MCD & 0.302 & 0.295  & \underline{1.035}  & 2.291 & 0.316 \\
  & DE & 0.204 & 0.205  & 1.886  & 2.034 & \underline{0.337} \\
  & HTS & 0.328 & 0.104  & 4.383  & 1.925 & 0.319 \\
  & HTS+MCD & 0.322 & \underline{0.103}  & 4.364  & \underline{1.924} & 0.314 \\
  & DUP & \underline{0.435} & 0.105  & 5.917  & 2.033 & 0.317 \\
\midrule
\multirow{6}{*}{\rotatebox{90}{\textsc{Xx-En}}}  
  & $\sigma^2$-fixed & -- & 0.067  & 3.200  & 1.831 & 0.080  \\
  & MCD & 0.274 & 0.204  & \underline{1.030}  & 1.965 & 0.079 \\
  & DE & 0.032 & 0.107 & 1.392 & 1.845 & \underline{0.111} \\
  & HTS & 0.246 & 0.069  & 2.755  & \underline{1.739} & 0.071 \\
  & HTS+MCD & 0.240 & 0.069  & 2.753  & \underline{1.739} & 0.068 \\
  & DUP & \underline{0.320} & \underline{0.064}  & 3.892  & 1.888 & 0.080 \\
 \midrule
 \multirow{6}{*}{\rotatebox{90}{\textsc{AVG}}}  
  & $\sigma^2$-fixed & -- &  0.086  & 3.910  & 1.903 & 0.192 \\
  & MCD & 0.287 & 0.247  & \underline{1.032}  & 2.120 & 0.191 \\
  & DE & 0.172 & 0.150 & 1.608  & 1.926 & \underline{0.210} \\
  & HTS & 0.285 & 0.086  & 3.527  & \underline{1.827} & 0.189 \\
  & HTS+MCD & 0.279 & 0.085  & 3.517  & \underline{1.827} & 0.185 \\
  & DUP & \underline{0.374} & \underline{0.083}  & 4.853  & 1.957 & 0.192 \\
\bottomrule
\end{tabular}
}
\addtolength{\tabcolsep}{-0.5pt}
\caption{Results for segment-level DA predictions by BLEURT  [\textbf{Average across language pairs}]. \underline{Underlined} numbers indicate the best result for each evaluation metric in each language pair.}
\label{tab:da_bleurt_all}
\end{table}

% En-Xx pairs
\begin{table}[ht]
\small
\centering
\addtolength{\tabcolsep}{-0.5pt}
\resizebox{7.7cm}{!}{
\begin{tabular}{clccccc}
\toprule
&  & UPS $\uparrow$ & ECE $\downarrow$ & Sha. $\downarrow$   & NLL $\downarrow$ & PPS $\uparrow$ \\  
\midrule
\multirow{6}{*}{\rotatebox{90}{\textsc{En-Cs}}}
  & $\sigma^2$-fixed & -- & 0.083  & 3.462  & 1.860 & 0.400 \\
  & MCD & 0.226 & 0.235  & \underline{1.031}  & 2.000 & 0.399\\
  & DE & 0.326 & 0.161  & 1.656  & \underline{1.846} & \underline{0.446} \\
  & HTS & 0.291 & \underline{0.071}  & 3.727  & 1.848 & 0.386 \\
  & HTS+MCD & 0.289 & \underline{0.071}  & 3.725  & 1.852 & 0.381 \\
  & DUP & \underline{0.410} & 0.079  & 5.029  & 1.986 & 0.400 \\
\midrule
\multirow{6}{*}{\rotatebox{90}{\textsc{En-De}}}
  & $\sigma^2$-fixed & -- & \underline{0.112}  & 5.779  & 2.099 & 0.257\\
  & MCD & 0.432 & {0.336}  & \underline{1.040}  & 2.603 & 0.256\\
  & DE & 0.198 & 0.217  & 2.508  & 2.207 & \underline{0.267} \\
  & HTS & 0.366 & 0.129  & 5.959  & 2.093 & 0.254 \\
  & HTS+MCD & 0.358 & 0.128  & 5.846  & \underline{2.083} & 0.248 \\
  & DUP & \underline{0.513} & 0.123  & 7.362  & 2.115 & 0.257 \\
\midrule
\multirow{6}{*}{\rotatebox{90}{\textsc{En-Ja}}} 
  & $\sigma^2$-fixed & -- & 0.134  & 6.040  & 2.105 & 0.277 \\
  & MCD & 0.326 & 0.358  & \underline{1.041}  & 2.592 & 0.275\\
  & DE & 0.121 & 0.258  & 1.937  & 2.185 & 0.289 \\
  & HTS & 0.278 & 0.126  & 5.105  & 2.003 & \underline{0.293} \\
  & HTS+MCD & 0.274 & 0.125  & 5.070  & \underline{2.000} & 0.287 \\
  & DUP & \underline{0.420} & \underline{0.124}  & 7.295  & 2.122 & 0.277 \\
\midrule
\multirow{6}{*}{\rotatebox{90}{\textsc{En-Pl}}}
  & $\sigma^2$-fixed & -- & 0.081  & 3.739  & 1.904 & 0.349\\
  & MCD & 0.217 & 0.247  & \underline{1.032}  & 2.105 & 0.347\\
  & DE & 0.247 & 0.154  & 1.728  & 1.876 & \underline{0.374} \\
  & HTS & 0.262 & 0.070  & 3.779  & \underline{1.870} & 0.341 \\
  & HTS+MCD & 0.260 & \underline{0.069}  & 3.739  & \underline{1.870} & 0.335 \\
  & DUP & \underline{0.362} & 0.076  & 5.092  & 1.992 & 0.349 \\
\midrule
\multirow{6}{*}{\rotatebox{90}{\textsc{En-Ru}}} 
  & $\sigma^2$-fixed & -- & 0.100  & 4.321  & 1.959 & 0.356 \\
  & MCD & 0.218 & 0.282  & \underline{1.034}  & 2.212 & 0.356 \\
  & DE & 0.158 & 0.184  & 1.838  & 1.951 & \underline{0.361} \\
  & HTS & 0.280 & 0.097  & 4.019  & 1.896 & 0.350 \\
  & HTS+MCD & 0.275 & \underline{0.096}  & 4.010  & \underline{1.895} & 0.345 \\
  & DUP & \underline{0.383} & 0.103  & 5.177  & 1.967 & 0.356 \\
\midrule
\multirow{6}{*}{\rotatebox{90}{\textsc{En-Ta}}} 
  & $\sigma^2$-fixed & -- & 0.065  & 3.530  & 1.877 & 0.240 \\
  & MCD & 0.131 & 0.218 & \underline{1.029} & 2.047 & 0.240 \\
  & DE & 0.162 & 0.135  & 1.822  & 1.864 & \underline{0.278} \\
  & HTS & \underline{0.320} & 0.062  & 3.499  & \underline{1.826} & 0.248 \\
  & HTS+MCD & 0.317 & 0.062  & 3.534  & 1.828 & 0.249 \\
  & DUP & 0.279 & \underline{0.060}  & 4.501  & 1.966 & 0.240 \\
\midrule
\multirow{6}{*}{\rotatebox{90}{\textsc{En-Zh}}} 
  & $\sigma^2$-fixed & -- & \underline{0.154}  & 6.063  & 2.097 & 0.280 \\
  & MCD & 0.506 & 0.375 & \underline{1.041} & 2.540 & 0.280 \\
  & DE & 0.171 & 0.297  & 1.940  & 2.297 & 0.282 \\
  & HTS & 0.465 & 0.160  & 4.919  & 1.975 & \underline{0.301} \\
  & HTS+MCD & 0.449 & 0.160  & 4.920  & \underline{1.974} & 0.297 \\
  & DUP & \underline{0.601} & \underline{0.154}  & 7.044  & 2.091 & 0.280 \\
\bottomrule
\end{tabular}
}
\addtolength{\tabcolsep}{-0.5pt}
\caption{Results for segment-level DA predictions by BLEURT for En-Xx LPs. \underline{Underlined} numbers indicate the best result for each evaluation metric in each language pair.}
\label{tab:da_bleurt_enxx}
\end{table}

% Xx-En pairs
\begin{table}[ht]
\small
\centering
\addtolength{\tabcolsep}{-0.5pt}
\resizebox{7.7cm}{!}{
\begin{tabular}{clccccc}
\toprule
&  & UPS $\uparrow$ & ECE $\downarrow$ & Sha. $\downarrow$   & NLL $\downarrow$ & PPS $\uparrow$ \\  
\midrule
\multirow{6}{*}{\rotatebox{90}{\textsc{Cs-En}}} 
  & $\sigma^2$-fixed & -- & \underline{0.068}  & 3.085  & 1.816 & 0.061 \\
  & MCD & 0.303 & 0.202  & \underline{1.030}  & 1.924 & 0.059 \\
  & DE & 0.020 & 0.156  & 1.271  & 1.806 & \underline{0.071} \\
  & HTS & 0.291 & 0.073  & 2.775  & 1.735 & 0.052 \\
  & HTS+MCD & 0.287 & 0.072  & 2.773  & \underline{1.734} & 0.050 \\
  & DUP & \underline{0.360} & \underline{0.068}  & 3.781  & 1.838 & 0.061 \\
\midrule
\multirow{6}{*}{\rotatebox{90}{\textsc{De-En}}}  
  & $\sigma^2$-fixed & -- & \underline{0.080}  & 3.053  & 1.808 & 0.018 \\
  & MCD & 0.364 & 0.210  & \underline{1.028}  & 1.908 & 0.016 \\
  & DE & 0.087 & 0.159  & 1.320  & 1.767 & \underline{0.044} \\
  & HTS & 0.326 & 0.091  & 3.015  & 1.748 & -0.002 \\
  & HTS+MCD & 0.323 & 0.091  & 3.000  & \underline{1.746} & -0.003 \\
  & DUP & \underline{0.399} & 0.081  & 3.981  & 1.837 & 0.018 \\
\midrule
\multirow{6}{*}{\rotatebox{90}{\textsc{Ja-En}}}  
  & $\sigma^2$-fixed & -- & 0.072  & 3.701  & 1.896 & 0.095\\
  & MCD & 0.228 & 0.229  & \underline{1.033}  & 2.082 & 0.094\\
  & DE & 0.011 & 0.143  & 1.524  & 1.845 & \underline{0.102} \\
  & HTS & 0.192 & 0.069  & 2.973  & 1.779 & 0.084 \\
  & HTS+MCD & 0.184 & 0.068  & 2.948  & \underline{1.777} & 0.082 \\
  & DUP & \underline{0.262} & \underline{0.066}  & 4.328  & 1.959 & 0.095 \\
\midrule
\multirow{6}{*}{\rotatebox{90}{\textsc{Km-En}}} 
  & $\sigma^2$-fixed & -- & 0.032  & 2.613  & 1.753 & 0.208 \\
  & MCD & 0.084 & 0.145 & \underline{1.029} & 1.827 & 0.205\\
  & DE & 0.020 & \underline{0.002} & 1.207 & 2.003 & \underline{0.278} \\
  & HTS & 0.100 & 0.026  & 2.168  & 1.642 & 0.209 \\
  & HTS+MCD & 0.099 & 0.026  & 2.152  & \underline{1.640} & 0.199 \\
  & DUP & \underline{0.149} & 0.021  & 3.222  & 1.856 & 0.208 \\
\midrule
\multirow{6}{*}{\rotatebox{90}{\textsc{Pl-En}}} 
  & $\sigma^2$-fixed & -- & 0.068  & 3.324  & 1.853 & 0.053 \\
  & MCD & 0.313 & 0.213 & \underline{1.031}  & 2.002 & 0.053\\
  & DE & 0.051 & 0.128  & 1.536  & 1.816 & \underline{0.065} \\
  & HTS & 0.253 & 0.069  & 2.842  & \underline{1.766} & 0.057 \\
  & HTS+MCD & 0.248 & 0.069  & 2.862  & 1.767 & 0.051 \\
  & DUP & \underline{0.358} & \underline{0.067}  & 3.922  & 1.919 & 0.053 \\
\midrule
\multirow{6}{*}{\rotatebox{90}{\textsc{Ps-En}}}  
  & $\sigma^2$-fixed & -- & 0.034  & 2.633  & 1.768 & 0.090 \\
  & MCD & 0.176 & 0.151 & \underline{1.029}  & 1.861 & 0.088 \\
  & DE & -0.029 & \underline{0.002}  & 1.267  & 2.134 & \underline{0.101} \\
  & HTS & 0.165 & 0.033  & 2.058  & \underline{1.647} & 0.075 \\
  & HTS+MCD & 0.159 & 0.033  & 2.044  & \underline{1.647} & 0.070 \\
  & DUP & \underline{0.218} & 0.028  & 3.267  & 1.859 & 0.090 \\
\midrule
\multirow{6}{*}{\rotatebox{90}{\textsc{Ru-En}}}  
  & $\sigma^2$-fixed & -- & 0.085  & 3.490  & 1.857 & 0.076 \\
  & MCD & 0.261 & 0.228 & \underline{1.030}  & 1.986 & 0.076 \\
  & DE & 0.027 & 0.153  & 1.482  & 1.815 & \underline{0.084} \\
  & HTS & 0.219 & 0.087  & 3.135  & 1.786 & 0.068 \\
  & HTS+MCD & 0.212 & 0.087  & 3.148  & \underline{1.785} & 0.067 \\
  & DUP & \underline{0.319} & \underline{0.081}  & 4.021  & 1.897 & 0.076 \\
\midrule
\multirow{6}{*}{\rotatebox{90}{\textsc{Ta-En}}}  
  & $\sigma^2$-fixed & -- & 0.041  & 2.069  & 1.658 & 0.108 \\
  & MCD & 0.238 & 0.121 & \underline{1.020}  & 1.688 & 0.106 \\
  & DE & 0.023 & 0.064  & 1.401  & 1.601 & \underline{0.132} \\
  & HTS & 0.245 & 0.046  & 1.704  & \underline{1.561} & 0.098 \\
  & HTS+MCD & 0.238 & 0.046  & 1.703  & 1.562 & 0.096 \\
  & DUP & \underline{0.273} & \underline{0.034}  & 2.959  & 1.767 & 0.108 \\
\midrule
\multirow{6}{*}{\rotatebox{90}{\textsc{Zh-En}}}  
  & $\sigma^2$-fixed & -- & 0.076  & 3.542  & 1.886 & 0.080 \\
  & MCD & 0.309 & 0.231 & \underline{1.033} & 2.075 & 0.079 \\
  & DE & 0.034 & 0.138  & 1.527  & 1.817 & \underline{0.097} \\
  & HTS & 0.278 & 0.078  & 2.984 & \underline{1.790} & 0.071 \\
  & HTS+MCD & 0.270 & 0.078  & 2.984  & \underline{1.790} & 0.068\\
  & DUP & \underline{0.353} & \underline{0.075}  & 4.252  & 1.932 & 0.080 \\
\bottomrule
\end{tabular}
}
\addtolength{\tabcolsep}{-0.5pt}
\caption{Results for segment-level DA predictions by BLEURT for Xx-En LPs. \underline{Underlined} numbers indicate the best result for each evaluation metric in each language pair.}
\label{tab:da_bleurt_xxen}
\end{table}

\begin{table}[!h]
\small
\centering
\addtolength{\tabcolsep}{-0.5pt}
\resizebox{7.7cm}{!}{
\begin{tabular}{clccccc}
\toprule
&  & UPS $\uparrow$ & ECE $\downarrow$ & Sha. $\downarrow$   & NLL $\downarrow$ & PPS $\uparrow$ \\  
\midrule
\multirow{6}{*}{\rotatebox{90}{\textsc{En-Xx}}}   & $\sigma^2$-fixed & -- & 0.018  & \underline{0.224}  & 0.913 & 0.650 \\
  & MCD & 0.022 & 0.011  & 0.260  & 0.853 & 0.603 \\
  & DE & 0.129 & 0.015  & 0.264  & 0.878 & 0.647 \\
  & HTS & \underline{0.183} & 0.013  & 0.319  & 0.892 & 0.630 \\
  & HTS+MCD & 0.175 & \underline{0.004}  & 0.272  & \underline{0.792} & 0.587 \\
  & DUP & 0.139 & 0.015  & 0.250  & 0.924 & \underline{0.650} \\
\midrule
\multirow{6}{*}{\rotatebox{90}{\textsc{Xx-En}}}   & $\sigma^2$-fixed & -- & 0.027  & \underline{0.498}  & 1.512 & 0.289\\
  & MCD & 0.059 & 0.016  & 0.523  & 1.352 & 0.271 \\
  & DE & 0.049 & 0.028  & 0.503  & 1.488 & \underline{0.293} \\
  & HTS & \underline{0.090} & 0.026  & 0.525  & 1.506 & 0.288 \\
  & HTS+MCD & 0.079 & \underline{0.010}  & 0.572  & \underline{1.351} & 0.267 \\
  & DUP & 0.065 & 0.026  & 0.518  & 1.546 & 0.289 \\
\midrule
\multirow{6}{*}{\rotatebox{90}{\textsc{AVG}}} 
& $\sigma^2$-fixed & -- & 0.023  & \underline{0.368}  & 1.228 & 0.460 \\
  & MCD & 0.041 & 0.014  & 0.398  & 1.115 & 0.428 \\
  & DE & 0.087 & 0.022  & 0.390  & 1.198 & \underline{0.461} \\
  & HTS & \underline{0.134} & 0.020  & 0.427  & 1.215 & 0.450 \\
  & HTS+MCD & 0.124 & \underline{0.007}  & 0.429  & \underline{1.085} & 0.419 \\
  & DUP & 0.100 & 0.021  & 0.391  & 1.251 & 0.460 \\
\bottomrule
\end{tabular}
}
\addtolength{\tabcolsep}{-0.5pt}
\caption{Results for segment-level DA predictions by UniTE [\textbf{Average across language pairs}]. \underline{Underlined} numbers indicate the best result for each evaluation metric in each language pair.}
\label{tab:da_unite_avg}
\end{table}

% En-Xx pairs
\begin{table}[!h]
\small
\centering
\addtolength{\tabcolsep}{-0.5pt}
\resizebox{7.7cm}{!}{
\begin{tabular}{clccccc}
\toprule
&  & UPS $\uparrow$ & ECE $\downarrow$ & Sha. $\downarrow$   & NLL $\downarrow$ & PPS $\uparrow$ \\  
\midrule
\multirow{6}{*}{\rotatebox{90}{\textsc{En-Cs}}}
  & $\sigma^2$-fixed & -- & 0.013  & 0.284  & 0.990 & \underline{0.735}\\
  & MCD & -0.007 & 0.010  & \underline{0.242}  & 0.803 & 0.672 \\
  & DE & 0.108 & 0.011  & 0.348  & 0.946 & 0.732 \\
  & HTS & \underline{0.151} & 0.009  & 0.427  & 0.961 & 0.718 \\
  & HTS+MCD & 0.132 & \underline{0.003}  & 0.249  & \underline{0.743} & 0.672 \\
  & DUP & 0.108 & 0.011  & 0.300  & 0.988 & \underline{0.735} \\
\midrule
\multirow{6}{*}{\rotatebox{90}{\textsc{En-De}}}  
  & $\sigma^2$-fixed & -- &  0.038  & 0.145  & 1.178 & \underline{0.623} \\
  & MCD & 0.036 & 0.017  & 0.227  & 0.890 & 0.579 \\
  & DE & 0.172 & 0.032  & \underline{0.191}  & 1.096 & \underline{0.623} \\
  & HTS & 0.262 & 0.030  & 0.281  & 1.064 & 0.603 \\
  & HTS+MCD & \underline{0.283} & \underline{0.008}  & 0.236  & \underline{0.798} & 0.578 \\
  & DUP & 0.241 & 0.031  & 0.207  & 1.213 & \underline{0.623} \\
\midrule
\multirow{6}{*}{\rotatebox{90}{\textsc{En-Ja}}}  
  & $\sigma^2$-fixed & -- &  0.011  & 0.160  & 0.718 & 0.688 \\
  & MCD & -0.015 & 0.008  & 0.241  & 0.775 & 0.650 \\
  & DE & 0.112 & 0.008  & 0.181  & \underline{0.659} & \underline{0.698} \\
  & HTS & 0.117 & 0.008  & 0.196  & 0.705 & 0.687 \\
  & HTS+MCD & \underline{0.148} & \underline{0.004}  & 0.229  & 0.684 & 0.646 \\
  & DUP & 0.089 & 0.008 & \underline{0.178}  & 0.728 & 0.688 \\
\midrule
\multirow{6}{*}{\rotatebox{90}{\textsc{En-Pl}}}  
  & $\sigma^2$-fixed & -- &  0.014  & 0.344  & 1.124 & 0.647 \\
  & MCD & -0.091 & 0.011  & \underline{0.299}  & 0.935 & 0.605 \\
  & DE & 0.121 & 0.011  & 0.426  & 1.074 & \underline{0.652} \\
  & HTS & 0.127 & 0.014  & 0.481  & 1.086 & 0.626 \\
  & HTS+MCD & \underline{0.141} & \underline{0.004}  & 0.345  & \underline{0.907} & 0.591 \\
  & DUP & 0.060 & 0.013  & 0.364  & 1.133 & 0.647 \\
\midrule
\multirow{6}{*}{\rotatebox{90}{\textsc{En-Ru}}} 
  & $\sigma^2$-fixed & -- &  0.026  & \underline{0.229}  & 1.001 & 0.610 \\
  & MCD & -0.059 & 0.015  & 0.248  & \underline{0.865} & 0.587 \\
  & DE & 0.112 & 0.020  & 0.284  & 1.017 & \underline{0.611} \\
  & HTS & 0.129 & 0.013  & 0.357  & 1.056 & 0.600 \\
  & HTS+MCD & \underline{0.148} & \underline{0.009}  & 0.273  & 0.890 & 0.576 \\
  & DUP & 0.085 & 0.022  & 0.256  & 1.040 & 0.610 \\
\midrule
\multirow{6}{*}{\rotatebox{90}{\textsc{En-Ta}}} 
  & $\sigma^2$-fixed & -- &  0.011  & \underline{0.334}  & 1.125 & 0.685\\
  & MCD & 0.079 & 0.005  & 0.396  & 1.052 & 0.647 \\
  & DE & 0.120 & 0.018  & 0.335  & 1.089 & \underline{0.688} \\
  & HTS & \underline{0.228} & 0.015  & 0.390  & 1.096 & 0.678 \\
  & HTS+MCD & 0.207 & \underline{0.003}  & 0.410  & \underline{1.032} & 0.634 \\
  & DUP & 0.164 & 0.009  & 0.371  & 1.110 & 0.685 \\
\midrule
\multirow{6}{*}{\rotatebox{90}{\textsc{En-Zh}}} 
  & $\sigma^2$-fixed & -- &  0.019  & \underline{0.101}  & 0.505 & \underline{0.571} \\
  & MCD & 0.177 & 0.011  & 0.214  & 0.760 & 0.504 \\
  & DE & 0.162 & 0.012  & 0.109  & \underline{0.486} & 0.546 \\
  & HTS & \underline{0.272} & 0.009  & 0.134  & 0.494 & 0.522 \\
  & HTS+MCD & 0.204 & \underline{0.002}  & 0.213  & 0.622 & 0.453 \\
  & DUP & 0.234 & 0.014  & 0.125  & 0.508 & \underline{0.571} \\
\bottomrule
\end{tabular}
}
\addtolength{\tabcolsep}{-0.5pt}
\caption{Results for segment-level DA predictions by UniTE for En-Xx LPs. \underline{Underlined} numbers indicate the best result for each evaluation metric in each language pair.}
\label{tab:da_unite_enxx}
\end{table}

% Xx-En pairs
\begin{table}[!h]
\small
\centering
\addtolength{\tabcolsep}{-0.5pt}
\resizebox{7.7cm}{!}{
\begin{tabular}{clccccc}
\toprule
&  & UPS $\uparrow$ & ECE $\downarrow$ & Sha. $\downarrow$   & NLL $\downarrow$ & PPS $\uparrow$ \\  
\midrule
\multirow{6}{*}{\rotatebox{90}{\textsc{Cs-En}}}  
  & $\sigma^2$-fixed & -- & 0.027  & 0.507  & 1.485 & 0.176  \\
  & MCD & 0.021 & 0.008  & 0.531  & 1.296 & 0.157 \\
  & DE & 0.016 & 0.029  & 0.491  & 1.454 & \underline{0.179} \\
  & HTS & \underline{0.049} & 0.029  & \underline{0.474}  & 1.462 & 0.171 \\
  & HTS+MCD & 0.037 & \underline{0.005}  & 0.563  & \underline{1.273} & 0.150 \\
  & DUP & 0.032 & 0.027  & 0.510  & 1.490 & 0.176 \\
\midrule
\multirow{6}{*}{\rotatebox{90}{\textsc{De-En}}}   
  & $\sigma^2$-fixed & -- & 0.037  & \underline{0.294}  & 1.492 & 0.551  \\
  & MCD & 0.109 & 0.018  & 0.337  & \underline{1.242} & 0.532 \\
  & DE & 0.089 & 0.033  & 0.320  & 1.437 & \underline{0.558} \\
  & HTS & 0.116 & 0.028  & 0.390  & 1.443 & 0.554 \\
  & HTS+MCD & \underline{0.117} & \underline{0.008}  & 0.434  & 1.247 & 0.526 \\
  & DUP & 0.093 & \underline{0.036}  & 0.303  & 1.532 & 0.551 \\
\midrule
\multirow{6}{*}{\rotatebox{90}{\textsc{Ja-En}}}   
  & $\sigma^2$-fixed & -- & 0.018  & 0.518  & 1.360 & 0.315  \\
  & MCD & 0.111 & \underline{0.005}  & 0.647  & 1.289 & 0.288 \\
  & DE & 0.073 & 0.022  & \underline{0.498}  & 1.366 & \underline{0.319} \\
  & HTS & \underline{0.128} & 0.018  & 0.547  & 1.425 & 0.310 \\
  & HTS+MCD & 0.116 & 0.008  & 0.669  & \underline{1.281} & 0.280 \\
  & DUP & 0.100 & 0.014  & 0.547  & 1.420 & 0.315 \\
\midrule
\multirow{6}{*}{\rotatebox{90}{\textsc{Km-En}}}  
  & $\sigma^2$-fixed & -- & 0.006  & 0.689  & \underline{1.251} & \underline{0.428}\\
  & MCD & 0.087 & 0.015  & 0.773  & 1.272 & 0.407 \\
  & DE & 0.040 & \underline{0.003}  & \underline{0.650}  &  1.253 & 0.425 \\
  & HTS & \underline{0.125} & \underline{0.003}  & 0.659  & 1.257 & 0.416 \\
  & HTS+MCD & 0.124 & 0.016  & 0.944  & 1.311 & 0.396 \\
  & DUP & 0.096 & 0.005  & 0.816  & 1.282 & \underline{0.428} \\
\midrule
\multirow{6}{*}{\rotatebox{90}{\textsc{Pl-En}}} 
  & $\sigma^2$-fixed & -- &  0.033  & 0.523  & 1.587 & 0.196  \\
  & MCD & 0.011 & 0.022  & \underline{0.484}  & \underline{1.408} & 0.178 \\
  & DE & 0.048 & 0.032  & 0.551  & 1.546 & \underline{0.202} \\
  & HTS & \underline{0.061} & 0.034 & 0.545  & 1.552 & 0.199 \\
  & HTS+MCD & 0.053 & \underline{0.012}  & 0.523  & 1.409 & 0.181 \\
  & DUP & 0.039 & 0.032  & 0.545  & 1.612 & 0.196 \\
\midrule
\multirow{6}{*}{\rotatebox{90}{\textsc{Ps-En}}}  
  & $\sigma^2$-fixed & -- & \underline{0.003}  & 0.792  & 1.350 & 0.260  \\
  & MCD & 0.064 & 0.011  & 0.861  & \underline{1.333} & 0.245 \\
  & DE & 0.022 & 0.004  & \underline{0.768}  & 1.342 & \underline{0.264} \\
  & HTS & \underline{0.080} & 0.005  & 0.772  & 1.353 & 0.259 \\
  & HTS+MCD & 0.062 & 0.012  & 0.970  & 1.364 & 0.243 \\
  & DUP & 0.068 & \underline{0.003}  & 0.803  & 1.349 & 0.260 \\
\midrule
\multirow{6}{*}{\rotatebox{90}{\textsc{Ru-En}}}  
  & $\sigma^2$-fixed & -- & 0.041  & \underline{0.368}  & 1.664 & 0.223  \\
  & MCD & 0.048 & 0.023  & 0.411  & 1.392 & 0.210 \\
  & DE & 0.064 & 0.041  & 0.385  & 1.658 & \underline{0.226} \\
  & HTS & \underline{0.102} & 0.040  & 0.404  & 1.649 & 0.225 \\
  & HTS+MCD & 0.088 & \underline{0.011}  & 0.455  & \underline{1.351} & 0.209 \\
  & DUP & 0.056 & 0.040  & 0.395  & 1.700 & 0.223 \\
\midrule
\multirow{6}{*}{\rotatebox{90}{\textsc{Ta-En}}}  
  & $\sigma^2$-fixed & -- &  0.031  & 0.607  & 1.481 & 0.322  \\
  & MCD & 0.038 & 0.022  & 0.609  & \underline{1.351} & 0.302 \\
  & DE & 0.035 & 0.033  & 0.605  & 1.470 & \underline{0.327} \\
  & HTS & \underline{0.073} & 0.026  & 0.648  & 1.568 & 0.320 \\
  & HTS+MCD & 0.065 & \underline{0.010}  & 0.627  & 1.357 & 0.298 \\
  & DUP & 0.061 & 0.030  & \underline{0.599}  & 1.592 & 0.322 \\
\midrule
\multirow{6}{*}{\rotatebox{90}{\textsc{Zh-En}}}  
  & $\sigma^2$-fixed & -- & 0.026  & 0.468  & 1.587 & 0.258  \\
  & MCD & 0.062 & 0.017  & \underline{0.455}  & \underline{1.417} & 0.242 \\
  & DE & 0.042 & 0.027  & 0.484  & 1.552 & \underline{0.262} \\
  & HTS & \underline{0.089} & 0.027  & 0.504  & 1.556 & 0.257 \\
  & HTS+MCD & 0.071 & \underline{0.011}  & 0.480  & 1.420 & 0.238 \\
  & DUP & 0.064 & 0.025  & 0.480  & 1.603 & 0.258 \\
\bottomrule
\end{tabular}
}
\addtolength{\tabcolsep}{-0.5pt}
\caption{Results for segment-level DA predictions by UniTE for Xx-En LPs. \underline{Underlined} numbers indicate the best result for each evaluation metric in each language pair.}
\label{tab:da_unite_xxen}
\end{table}

% \subsection{Bottleneck size for DUP}

% As mentioned in \S\ref{sec:dup_description}, DUP uses the same architecture with COMET, with the insertion of a bottleneck layer 

\end{document}